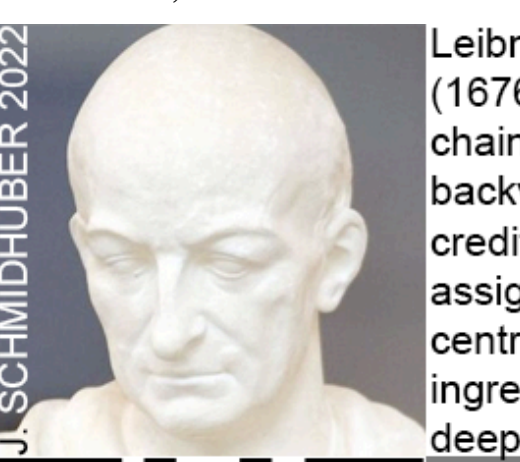

Leibniz (1676): chain rule for backward credit assignment, central ingredient of deep learning

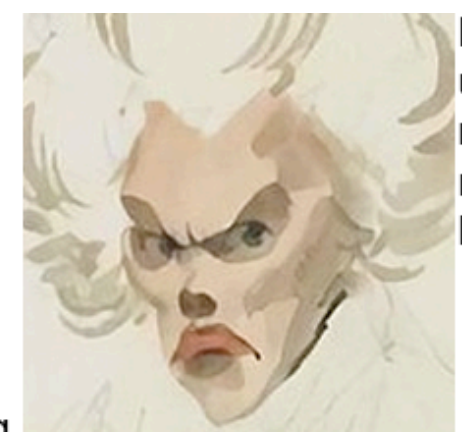

Legendre (1805) and Gauss (1795, unpublished): first linear neural networks (NNs) / linear regression / method of least squares / shallow learning

Famous example of pattern recognition through shallow learning from astronomical data: re-discovery of dwarf planet Ceres (Gauss, 1801)

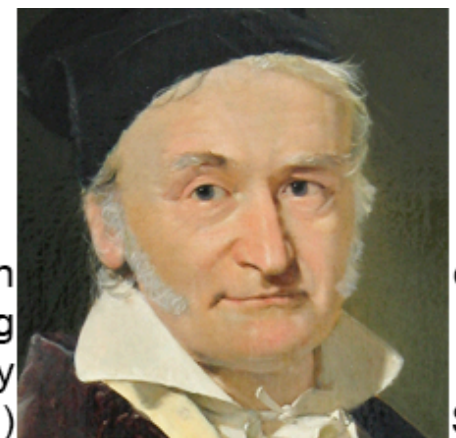

Cauchy (1847): gradient descent (GD), basic tool of deep learning. Robbins & Monro (1952): Stochastic GD

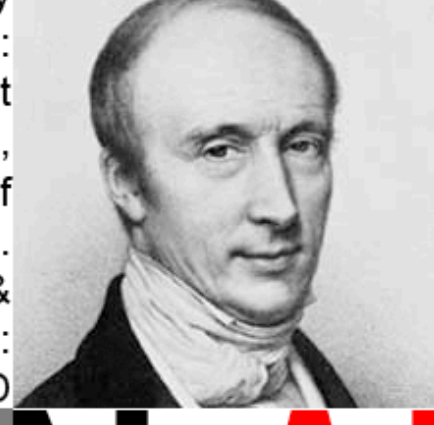

THE ROAD TO MODERN AI

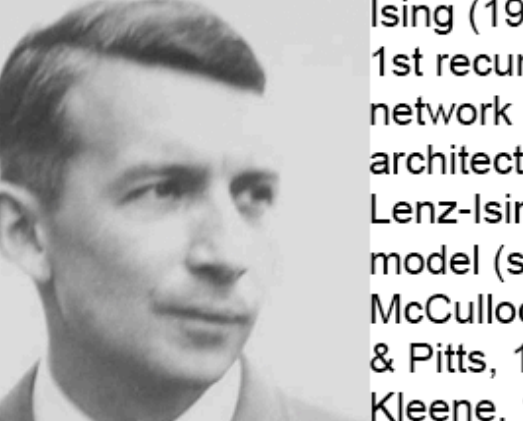

Ising (1925): 1st recurrent network architecture: Lenz-Ising model (see also McCulloch & Pitts, 1943, Kleene, 1956)

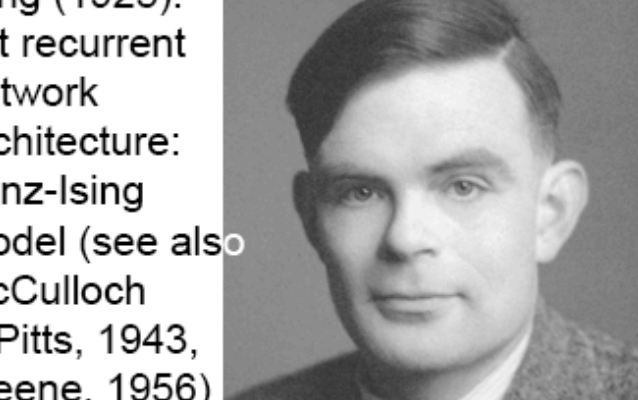

Turing (1948): unpublished ideas related to evolving recurrent NNs (RNNs)

Rosenblatt (1958): multilayer perceptron (MLP) (only last layer learned: no deep learning yet) See also Steinbuch (1961) Joseph (1961)

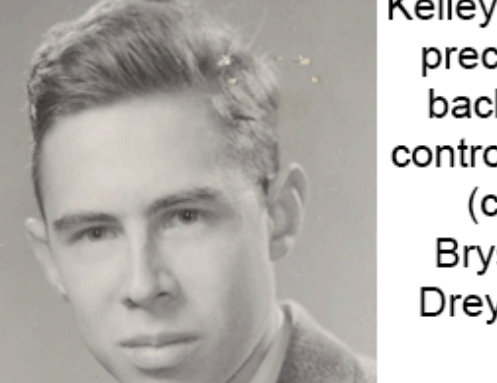

Kelley (1960): precursor of backprop in control theory (compare Bryson,'61; Dreyfus,'62)

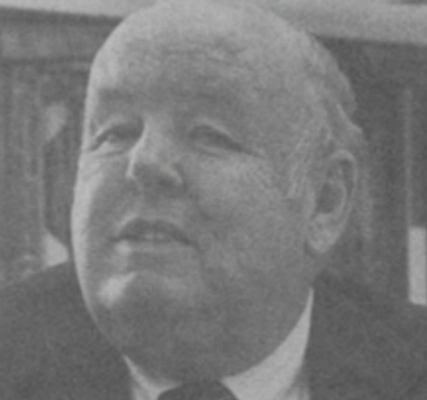

ARTIFICIAL NEURAL NETWORKS UP TO 1979

FROM SHALLOW LEARNING CIRCA 1800 TO DEEP LEARNING

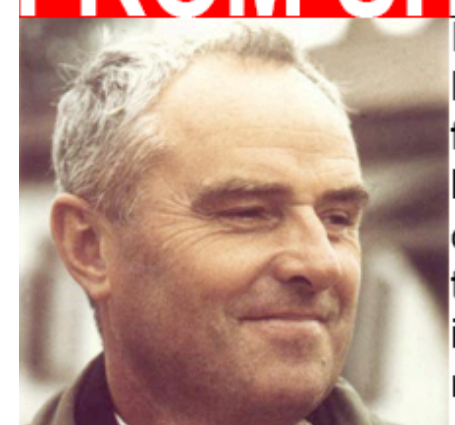

Ivakhnenko & Lapa (1965): first deep learning in deep MLPs that learn internal representations of input data

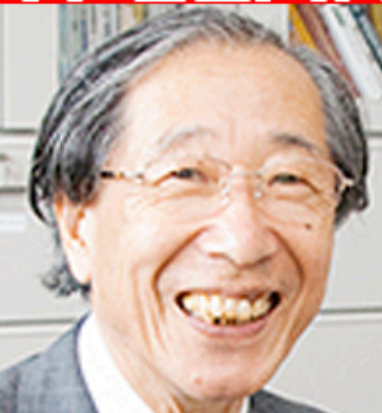

Amari (1967-68): deep learning by stochastic gradient descent for deep MLPs 1972: 1st published learning RNN based on Ising model (1925)

Linnainmaa (1970): backpropagation or reverse mode of automatic differentiation

First applied to NNs by Werbos (1982)

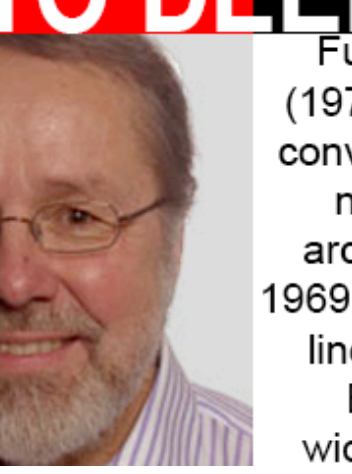

Fukushima (1979): deep convolutional neural net architecture 1969: rectified linear units. Both now widely used

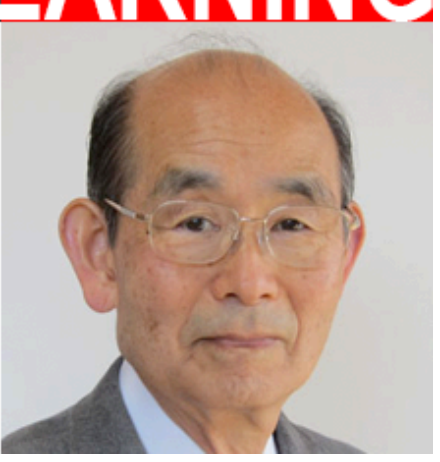

Jürgen Schmidhuber, KAUST GenAI, Swiss AI Lab IDSIA
Pronounce: You_again Shmidhoobuh

AI Blog
@SchmidhuberAI
juergen@idsia.ch

---

# Annotated History of Modern AI and Deep Learning

---

**Abstract.** Machine learning (ML) is the science of credit assignment. It seeks to find patterns in observations that explain and predict the consequences of events and actions. This then helps to improve future performance. Minsky's so-called *"fundamental credit assignment problem"* (1963) surfaces in all sciences including physics (why is the world the way it is?) and history (which persons/ideas/actions have shaped society and civilisation?). Here I focus on the history of ML itself. Modern artificial intelligence (AI) is dominated by artificial neural networks (NNs) and deep learning,[DL1-4] both of which are conceptually closer to the old field of cybernetics than what was traditionally called AI (e.g., expert systems and logic programming). A modern history of AI & ML must emphasize breakthroughs outside the scope of shallow AI text books. In particular, it must cover the mathematical foundations of today's NNs such as the chain rule (1676), the first NNs (circa 1800), the first practical AI (1914), the theory of AI and its limitations (1931-34), and the first working deep learning algorithms (1965-). From the perspective of 2025, I provide a timeline of the most significant events in the history of NNs, ML, deep learning, AI, computer science, and mathematics in general, crediting the individuals who laid the field's foundations. The text contains numerous hyperlinks to relevant overview sites from the AI Blog. It also debunks certain popular yet misleading historical accounts of AI and deep learning and—with a ten-year delay—supplements my 2015 award-winning deep learning survey[DL1] which provides hundreds of additional references. Finally, I will put things in a broader historical context, spanning from the Big Bang to when the universe will be many times older than it is now.

**Disclaimer.** Some say a history of deep learning should not be written by someone who has helped to shape it—*"you are part of history not a historian."*[CONN21] I cannot subscribe to that point of view. Since I seem to know more about deep learning history than others—and evidently much more than many who have tried to summarize the history of deep learning before,[S20][DL3,DL3a][DL1-2][DLP][NOB] I consider it my duty to document and promote this knowledge, even if that seems to imply a conflict of interest, as it means prominently mentioning my own team's work, because (as of 2025) the most cited NNs are based on it.[MOST] I leave it to future AI historians to correct any era-specific potential bias.

---

## Table of Contents

---

---

## Introduction

---

Over time, certain historic events have become more important in the eyes of certain beholders. For example, the Big Bang of 13.8 billion years ago is now widely considered an essential moment in the history of everything. Until a few decades ago, however, it has remained completely unknown to earthlings, who for a long time have entertained quite erroneous ideas about the origins of the universe (see the final section for more on the world's history). Currently accepted histories of many more limited subjects are results of similarly radical revisions. Here I will focus on the history of artificial intelligence (AI), which also isn't quite what it used to be.

A history of AI written in the 1980s would have emphasized topics such as theorem proving,[GOD][GOD34][ZU48][NS56] logic programming, expert systems, and heuristic search.[FEI63,83][LEN83] This would be in line with topics of a 1956 conference in Dartmouth, where the term "AI" was coined by John McCarthy as a way of describing an old area of research seeing renewed interest.

However, *practical AI* existed long before 1956, dating back at least to 1914, when Leonardo Torres y Quevedo (see below) built the first working chess end game player[BRU1-4] (back then chess was considered as an activity restricted to the realms of intelligent creatures).

Similar for *AI Theory*, which dates back at least to 1931-34 when Kurt Gödel (see below) identified fundamental limits of any type of computation-based AI.[GOD][BIB3][GOD21,a,b]

A history of AI written in the early 2000s would have put more emphasis on topics such as support vector machines and kernel methods,[SVM1-6] Bayesian (actually Laplacian or possibly Saundersonian[STI83-85]) reasoning[BAY1-8][FI22] and other concepts of probability theory and statistics,[MM1-5][NIL98][RUS95] decision trees,[e.g.,[MIT97]] ensemble methods,[ENS1-4] swarm intelligence,[SW1] and evolutionary computation.[EVO1-7]([TUR1],unpublished) Why? Because back then such techniques drove many successful AI applications.

A history of AI written in the 2020s must emphasize concepts such as the even older chain rule[LEI07] and deep nonlinear artificial neural networks (NNs) trained by gradient descent,[GD'] in particular, feedback-based recurrent networks, which are general computers whose programs are weight matrices.[AC90] Why? Because many of the most famous and most commercial recent AI applications depend on them.[DL4][GPT3]

Such NN concepts are actually conceptually close to topics of the MACY conferences (1946-1953)[MACY51] and the *1951 Paris conference on calculating machines and human thought*, now often viewed as the first conference on AI.[AI51][BRO21][BRU4] However, before 1956, much of what's now called AI was still called *cybernetics*, with a focus very much in line with modern AI based on "deep learning" with NNs.[DL1-2][DEC]

Although modern NNs date back to the late 1700s when people did not even know about biological neurons (see Sec. 3), some of the more recent NN research was inspired by the human brain, which has on the order of 100 billion neurons, each connected to 10,000 other neurons on average. Some are input neurons that feed the rest with data (sound, vision, tactile, pain, hunger). Others are output neurons that control muscles. Most neurons are hidden in between, where thinking takes place. Your brain apparently learns by changing the strengths or weights of the connections, which determine how strongly neurons influence each other, and which seem to encode all your lifelong experience. Similar for our *artifical* NNs, which learn better than previous methods to recognize speech or handwriting or video, minimize pain, maximize pleasure, drive cars, etc.[MIR](Sec. 0)[DL1-4]

How can NNs learn all of this? In what follows, I shall highlight essential historic contributions that made this possible. Since virtually all of the fundamental concepts of modern AI were derived in previous millennia (including the basics of Large Language Models such as ChatGPT), the section titles below emphasize developments only up to the year 2000. However, many of the sections mention the later impact of this work in the new millennium, which brought numerous improvements in hardware and software, a bit like the 20th century brought numerous improvements of the cars invented in the 19th.

The present piece also debunks a frequently repeated, misleading "history of deep learning"[S20][DL3,3a] which ignores most of the pioneering work mentioned below.[T22][DLP][NOB] See Footnote 5. The title image of the present article is a reaction to an erroneous piece of common knowledge which says[T19] that the use of NNs *"as a tool to help computers recognize patterns and simulate human intelligence had been introduced in the 1980s,"* although such NNs appeared long before the 1980s.[T22][DLP][NOB] Ensuring proper credit assignment in all of science is of great importance to me—just as it should be to all scientists—and I encourage an interested reader to also take a look at some of my letters on this in *Science* and *Nature*, e.g., on the history of aviation,[NASC1-2] the telephone,[NASC3] the computer,[NASC4-7] resilient robots,[NASC8] and scientists of the 19th century.[NASC9]

Finally, to round it off, I'll put things in a broader historic context spanning the time since the Big Bang until when the universe will be many times older than it is now.

---

# 1676: The Chain Rule For Backward Credit Assignment

---

In 1676, Gottfried Wilhelm Leibniz published the chain rule of differential calculus in a memoir (albeit with a sign error of all things!); Guillaume de l'Hospital described it in his 1696 textbook on Leibniz' differential calculus.[LEI07-10][L84] Today, this rule is central for credit assignment in deep neural networks (NNs). Why? The most popular NNs have nodes or neurons that compute differentiable functions of inputs from other neurons, which in turn compute differentiable functions of inputs from other neurons, and so on. The question is: how will the output of the final function change if we modify the parameters or weights of an earlier function a bit? The chain rule is the basic tool for computing the answer.

LEIBNIZ

This answer is used by the technique of *gradient descent* (GD), apparently first proposed by Augustin-Louis Cauchy in 1847[GD'] (and much later by Jacques Hadamard[GD'']; the stochastic version called SGD is due to Herbert Robbins and Sutton Monro (1951)[STO51-52]). To teach an NN to translate input patterns from a training set into desired output patterns, all NN weights are iteratively changed a bit in the direction of the biggest local improvement, to create a slightly better NN, and so on, until a satisfactory solution is found.

*Footnote 1.* In 1684, Leibniz was also the first to publish "modern" calculus;[L84][SON18][MAD05][LEI21,a,b] later Isaac Newton was also credited for his unpublished work.[SON18] Their priority dispute,[SON18] however, did *not* encompass the chain rule.[LEI07-10] Of course, both were building on earlier work: in the 2nd century B.C., Archimedes (sometimes called the greatest scientist ever[ARC06]) paved the way for *infinitesimals* and published special cases of calculus, e.g., for spheres and parabola segments, building on even earlier work in ancient Greece. Fundamental work on calculus was also conducted in the 14th century by Madhava of Sangamagrama and colleagues of the Indian Kerala school.[MAD86-05]

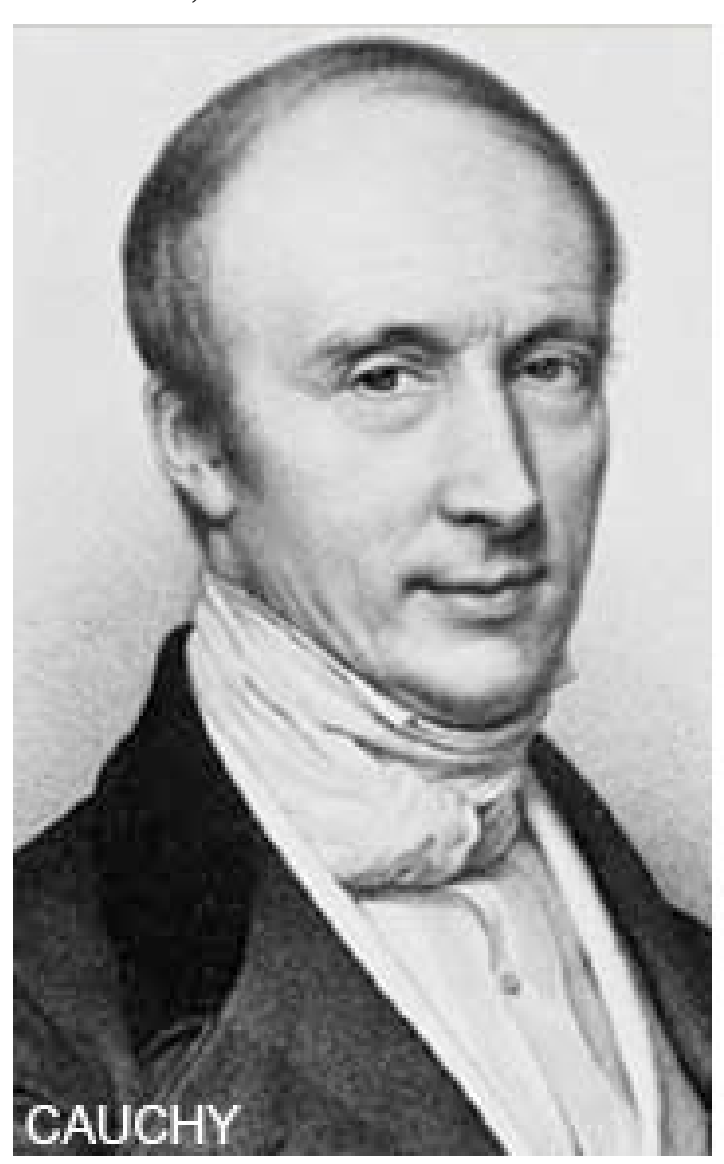

*Footnote 2.* Remarkably, Leibniz (1646-1714, *aka* "the world's first computer scientist"[LA14]) also laid foundations of modern computer science. He designed the first machine that could perform all four arithmetic operations (1673), and the first with an *internal memory*.[BL16] He described the principles of *binary computers* (1679)[L79][L03][LA14][HO66][LEI21,a,b] employed by virtually all modern machines. His formal *Algebra of Thought* (1686)[L86][WI48] was deductively equivalent[LE18] to the much later *Boolean Algebra* (1847).[BOO] His *Characteristica Universalis & Calculus Ratiocinator* aimed at answering all possible questions through computation;[WI48] his *"Calculemus!"* is one of the defining quotes of the age of enlightenment. It is quite remarkable that he is *also* responsible for the chain rule, foundation of "modern" deep learning, a key subfield of modern computer science.

*Footnote 3.* Some claim that the backpropagation algorithm (discussed further down; now widely used to train deep NNs) is just the chain rule of Leibniz (1676) popularised by L'Hospital (1696).[CONN21] No, it is the efficient way of applying the chain rule to big networks with differentiable nodes (there are also many inefficient ways of doing this).[BP4][T22][DLP][NOB] It was not published until 1970, as discussed below.[BP1,4,5]

---

# ~1800: First NN / Linear Regression / Shallow Learning

---

In 1805, Adrien-Marie Legendre published what's now called a 2-layer linear neural network (NN). Johann Carl Friedrich Gauss was credited for earlier unpublished work on this done circa 1795.[STI81] Back then, compute was many trillions of times more expensive than in 2025.

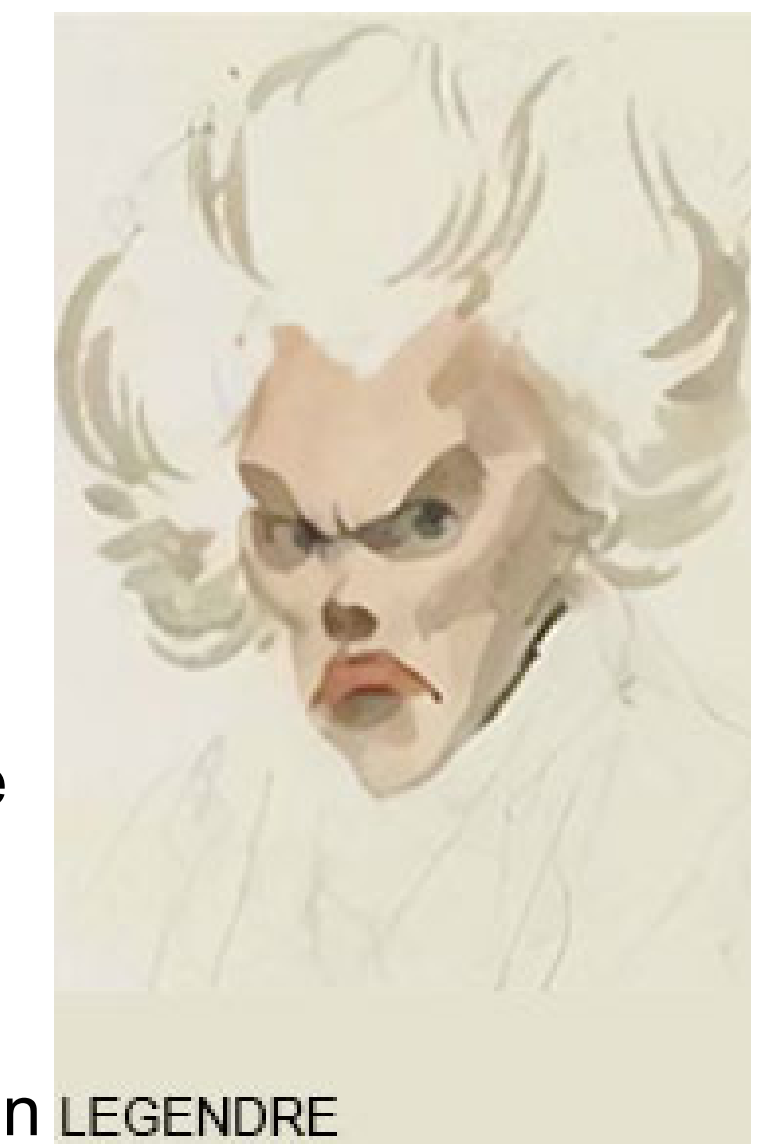

The Gauss-Legendre NN from over 2 centuries ago[NN25] has an input layer with several input units, and an output layer. For simplicity, let's assume the latter consists of a single output unit. Each input unit can hold a real-valued number and is connected to the output unit by a connection with a real-valued weight. The NN's output is the sum of the products of the inputs and their weights. Given a training set of input vectors and desired target values for each of them, the NN weights are adjusted such that the *sum of the squared errors* between the NN outputs and the corresponding targets is minimized.[DLH] Now the NN can be used to process previously unseen test data.

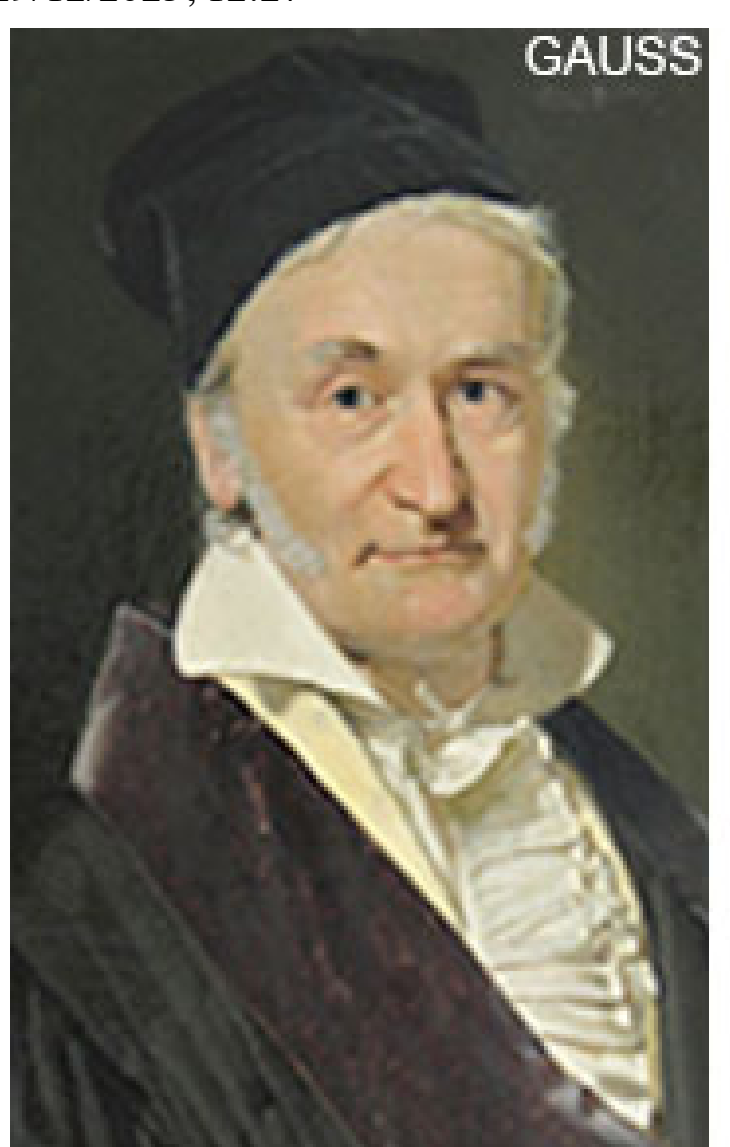

Of course, back then this was not called an NN, because people didn't even know about biological neurons yet (the first microscopic image of a nerve cell was created decades later by Valentin in 1836[CAJ06]). Instead, the technique was called *the Method of Least Squares*, also widely known in statistics as *Linear Regression*. But it is *mathematically identical* to today's linear 2-layer NNs: *same* basic algorithm, *same* error function, *same* adaptive parameters/weights. Such simple NNs perform *"shallow learning"* (as opposed to *"deep learning"* with many nonlinear layers[DL25]). In fact, many modern NN courses start by introducing this method, then move on to more complex, deeper NNs.[DLH]

Even the applications of the early 1800s were similar to today's: learn to predict the next element of a sequence, given previous elements. *That's what ChatGPT does!* The first famous example of pattern recognition through an NN dates back over 200 years: the rediscovery of the dwarf planet Ceres in 1801 through Gauss, who collected noisy data points from previous astronomical observations, then used them to adjust the parameters of a predictor, which essentially learned to generalise from the training data to correctly predict the new location of Ceres. That's what made the young Gauss famous.[DLH][NN25]

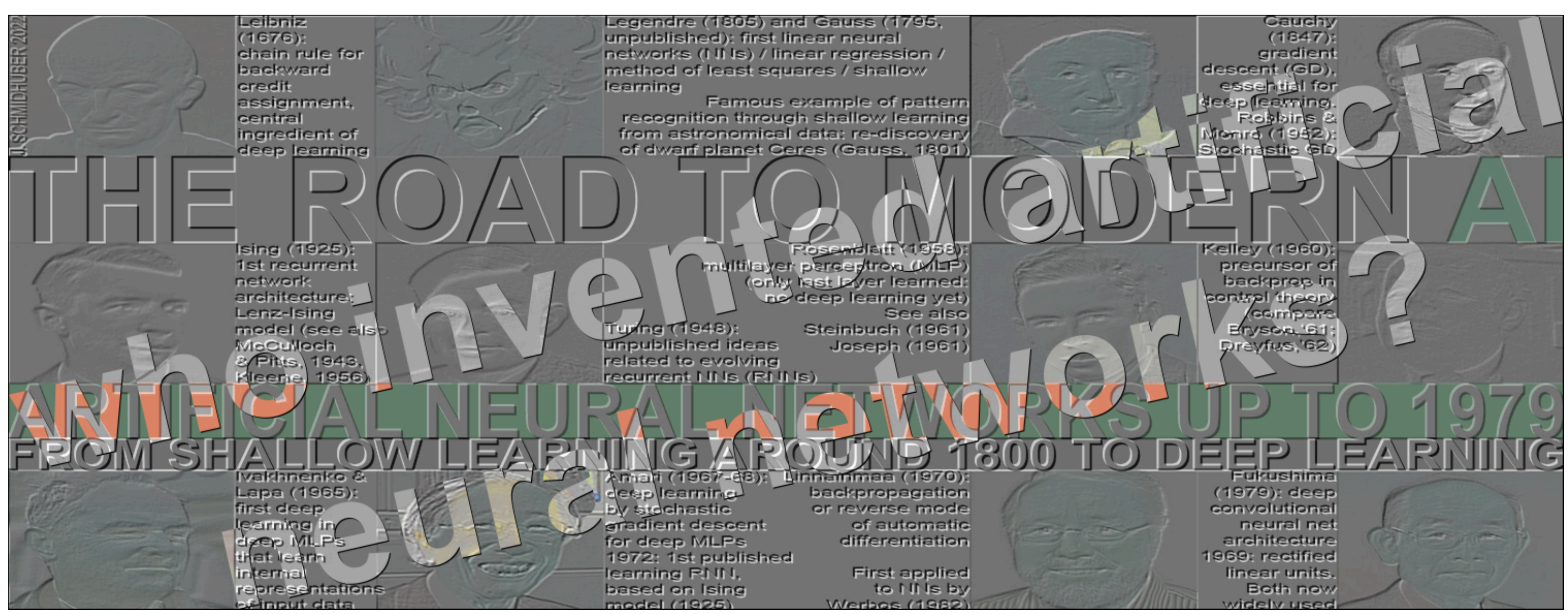

Some people believe that modern NNs were somehow inspired by the biological brain. But that's not the case: decades before biological nerve cells were discovered (1836) and the term *"neuron"* was coined by Waldeyer (1891),[CAJ06] plain engineering and mathematical problem solving already led to what's now called NNs. True, the *terminology* of artificial neural nets was introduced only much later in the 1900s. For example, certain non-learning NNs were discussed by McCulloch & Pitts in 1943.[MC43] Informal thoughts about a simple NN learning rule were published by Konorski in 1948[HEB48] and by Hebb in 1949.[HEB49] Evolutionary computation[EVO1-7] for NNs[EVONN1-3] was mentioned in Turing's unpublished 1948 report.[TUR1] Rosenblatt's perceptron (1958)[R58] combined a linear Gauss-Legendre NN (1795-1805) with an output threshold function to obtain a pattern classifier (compare his more advanced work on multi-layer networks discussed below). Joseph[R61] mentions an even earlier perceptron-like device by Farley & Clark (see also their earlier work[FAC54-55]). Widrow & Hoff's similar Adaline learned in 1962.[WID62] See also Selfridge's 1959 Pandemonium.[SE59] However, while these NN papers of the mid 1900s are of historical interest, *they have actually less to do with modern AI than the much older adaptive NN by Gauss & Legendre*, still heavily used in the 1990s and 2000s, the very foundation of all NNs, including the recent deeper NNs.[DLH][NN25] Remarkably, in the past 2 centuries, not so much has changed in AI research: as of 2025, NN progress is still mostly driven by engineering, not by neurophysiological insights. (Exceptions dating back many decades e.g.,[CN25] confirm the rule.)

Astonishingly, NN authors of the mid 1900s e.g.,[R58][R61] seemed unaware of the much earlier NN (1795-1805) famously known in the field of statistics as "method of least squares" or "linear regression." Remarkably, today's most frequently used 2-layer NNs are those of Gauss & Legendre, not those of the 1940s[MC43] and 1950s[R58] (which were not even differentiable)! See also: who invented artificial neural networks?[NN25]

*Footnote 4.* Today, students of all technical disciplines are required to take math classes, in particular, analysis, linear algebra, and statistics. In all of these fields, essential results and methods are (at least partially) due to Gauss: the fundamental theorem of algebra, Gauss elimination, the Gaussian distribution of statistics, etc. The so-called "greatest mathematician since antiquity" also pioneered differential geometry, number theory (his favorite subject), and non-Euclidean geometry. Furthermore, he made major contributions to astronomy and physics. Modern engineering including AI would be unthinkable without his results.

---

# 1920-1925: First Recurrent Network Architecture

---

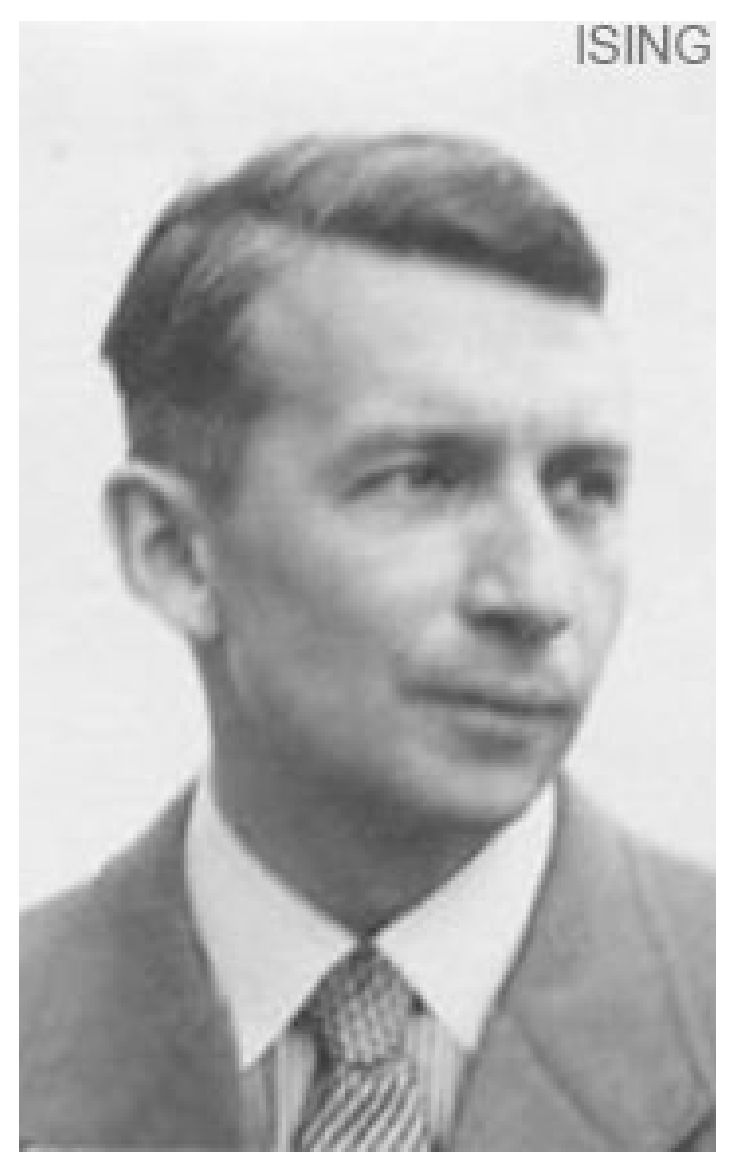

Like the human brain, but unlike the more limited *feedforward* NNs (FNNs), recurrent NNs (RNNs) have feedback connections, such that one can follow directed connections from certain internal nodes to others and eventually end up where one started. This is essential for implementing a memory of past events during sequence processing.

The first *non-learning* RNN architecture (the Ising model or Lenz-Ising model) was introduced and analyzed by physicists Ernst Ising and Wilhelm Lenz in the 1920s.[L20][I24,I25][K41][W45][T22][NOB] It settles into an equilibrium state in response to input conditions, and is the foundation of the first *learning* RNNs (see below).

Non-learning RNNs were also discussed in 1943 by neuroscientists Warren McCulloch und Walter Pitts[MC43] and formally analyzed in 1956 by Stephen Cole Kleene.[K56]

## ~1972: First Published Learning Artificial RNNs

In 1972, Shun-Ichi Amari made the Lenz-Ising recurrent architecture *adaptive* such that it could learn to associate input patterns with output patterns by changing its connection weights.[AMH1] See also Stephen Grossberg's work on biological networks,[GRO69] David Marr's[MAR71] and Teuvo Kohonen's[KOH72] work, and Kaoru Nakano's learning RNN.[NAK72]

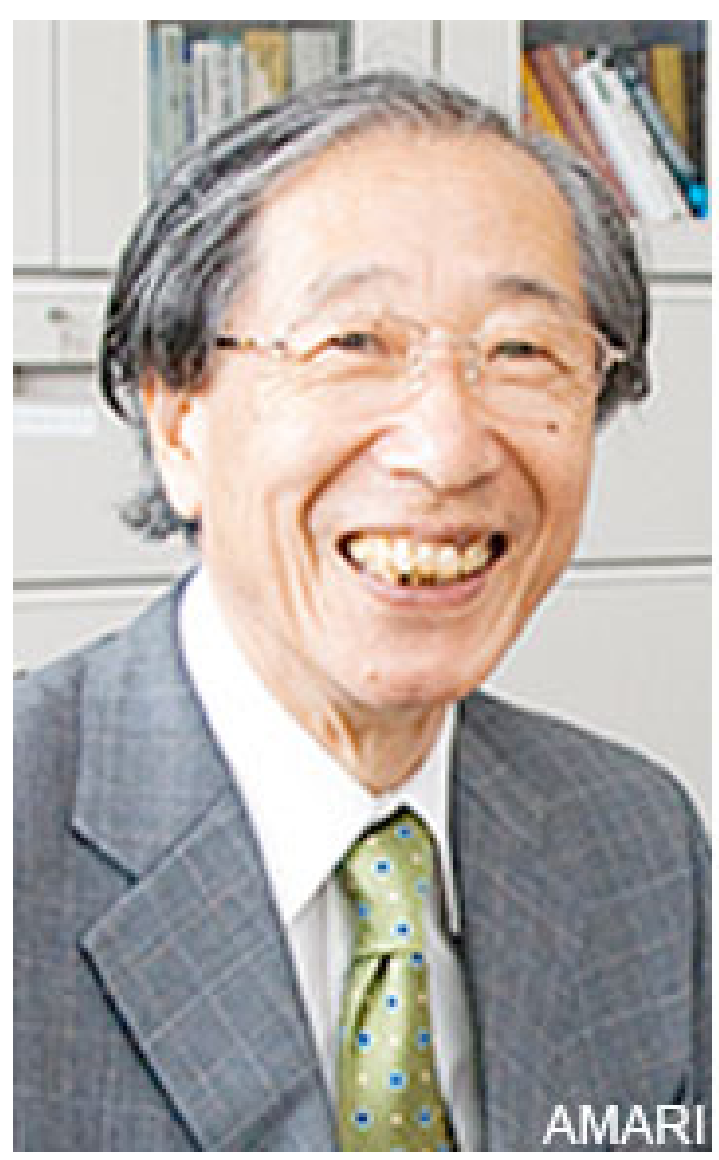

10 years later, the basic equations of the Amari network were republished (and its storage capacity analyzed).[AMH2][NOB] Some called it the Hopfield Network (!) or Amari-Hopfield Network.[AMH3] It does not process sequences but settles into an equilibrium in response to static input patterns. However, Amari (1972) also had a sequence-processing generalization thereof.[AMH1]

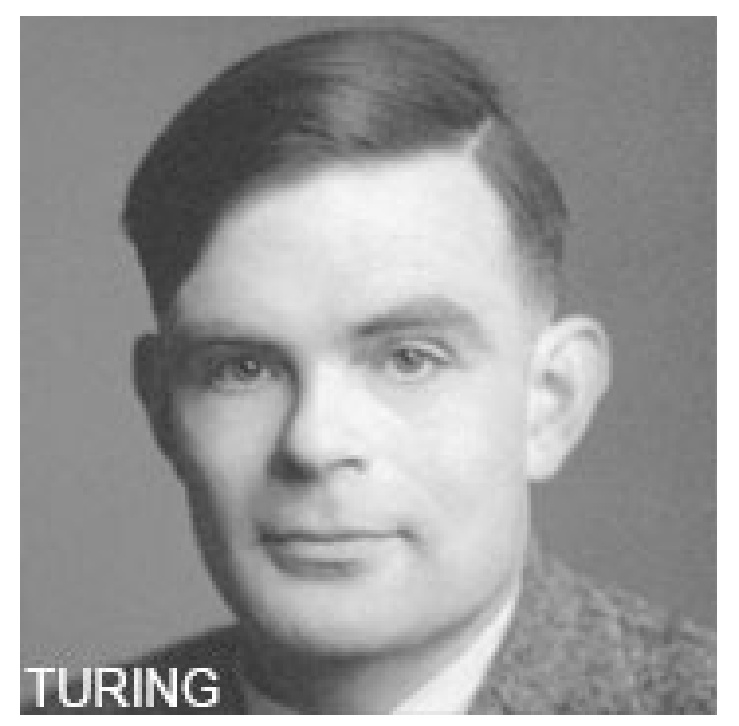

Remarkably, already in 1948, Alan Turing wrote up ideas related to artificial evolution and learning RNNs. This, however, was first published many decades later,[TUR1] which explains the obscurity of his thoughts here.[TUR21] (Margin note: it has been pointed out that the famous "Turing Test" should actually be called the "Descartes Test."[TUR3,a,b][TUR21])

Today, the most popular RNN is the Long Short-Term Memory (LSTM) mentioned below, which has become the most cited AI of the 20th century.[MOST]

---

# 1958: Multilayer Feedforward NN (without Deep Learning)

---

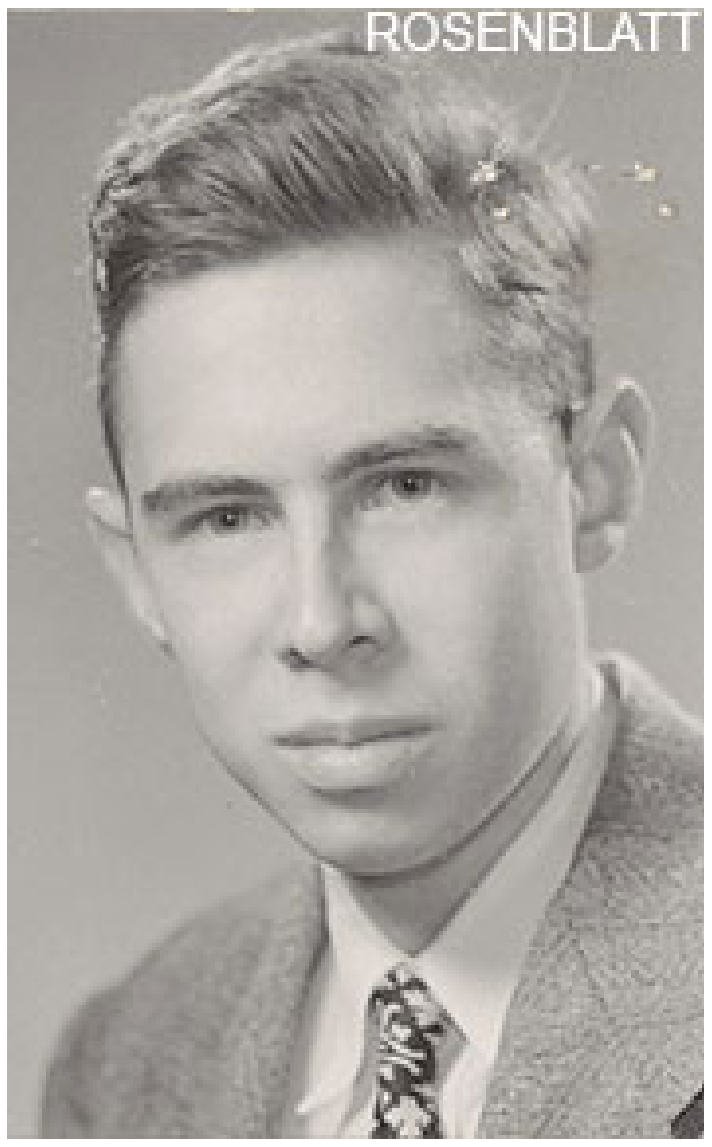

In 1958, Frank Rosenblatt not only combined linear NNs and threshold functions (see the section on shallow learning since 1800), he also had more interesting, deeper *multilayer* perceptrons (MLPs).[R58] His MLPs had a non-learning first layer with randomized weights and an adaptive output layer. Although this was not yet deep learning, because only the last layer learned,[DL1] Rosenblatt basically had what much later was rebranded as *Extreme Learning Machines (ELMs)* without proper attribution.[ELM1-2][CONN21][T22][DLP]

MLPs were also discussed in 1961 by Karl Steinbuch[ST61-95] and Roger David Joseph[R61] (1961). See also Oliver Selfridge's multilayer Pandemonium[SE59] (1959).

Following Joseph's 1961 preliminary ideas about training hidden units,[R61] Rosenblatt (1962) even wrote about *"back-propagating errors"* in an MLP with a hidden layer,[R62] but Joseph & Rosenblatt had no working *deep learning* algorithm for deep MLPs. What's now called backpropagation is quite different and was first published in 1970, as discussed below.[BP1-BP5][BPA-C]

Today, the most popular FNN is a variant of the LSTM-based Highway Net (mentioned below) called ResNet,[HW1-3][HW25,b] which has become the most cited AI of the 21st century.[MOST]

# 1965: First Deep Learning

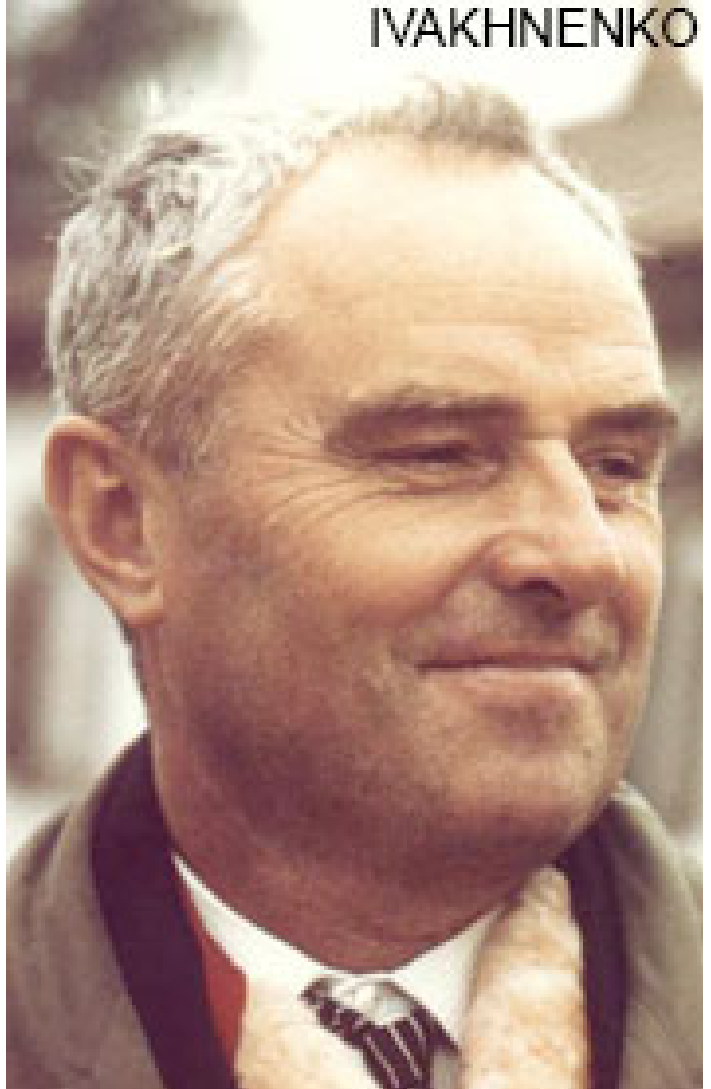

Successful learning in *deep* feedforward network architectures started in 1965 in Ukraine (back then the USSR) when Alexey Ivakhnenko & Valentin Lapa introduced the first general, working learning algorithms for deep multi-layer perceptrons (MLPs) or feedforward NNs (FNNs) with many hidden layers (already containing the now popular multiplicative gates).[DEEP1-2][DL1-2][DLH][DL25]

A paper of 1971[DEEP2] described a deep learning net with 8 layers, trained by their highly cited method which was still popular in the new millennium,[DL2] especially in Eastern Europe.[MIR](Sec. 1)[R8]

Given a training set of input vectors with corresponding target output vectors, layers are incrementally grown and trained by regression analysis. In a fine-tuning phase, superfluous hidden units are pruned through regularisation with the help of a separate validation set.[DEEP2][DLH] The numbers of layers and units per layer are learned in problem-dependent fashion. This is a generalization of the original 2-layer Gauss-Legendre NN (1795-1805).[DLP] See also: who invented artificial neural networks?[NN25]

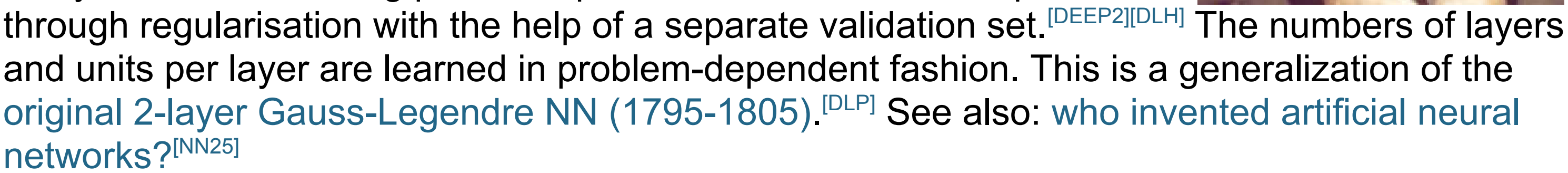

That is, *Ivakhnenko and colleagues had connectionism with adaptive hidden layers two decades before the name "connectionism" became popular in the 1980s*. Like later deep NNs, his nets learned to create hierarchical, distributed, *internal representations* of incoming data. He did not call them *deep learning* NNs, but that's what they were.

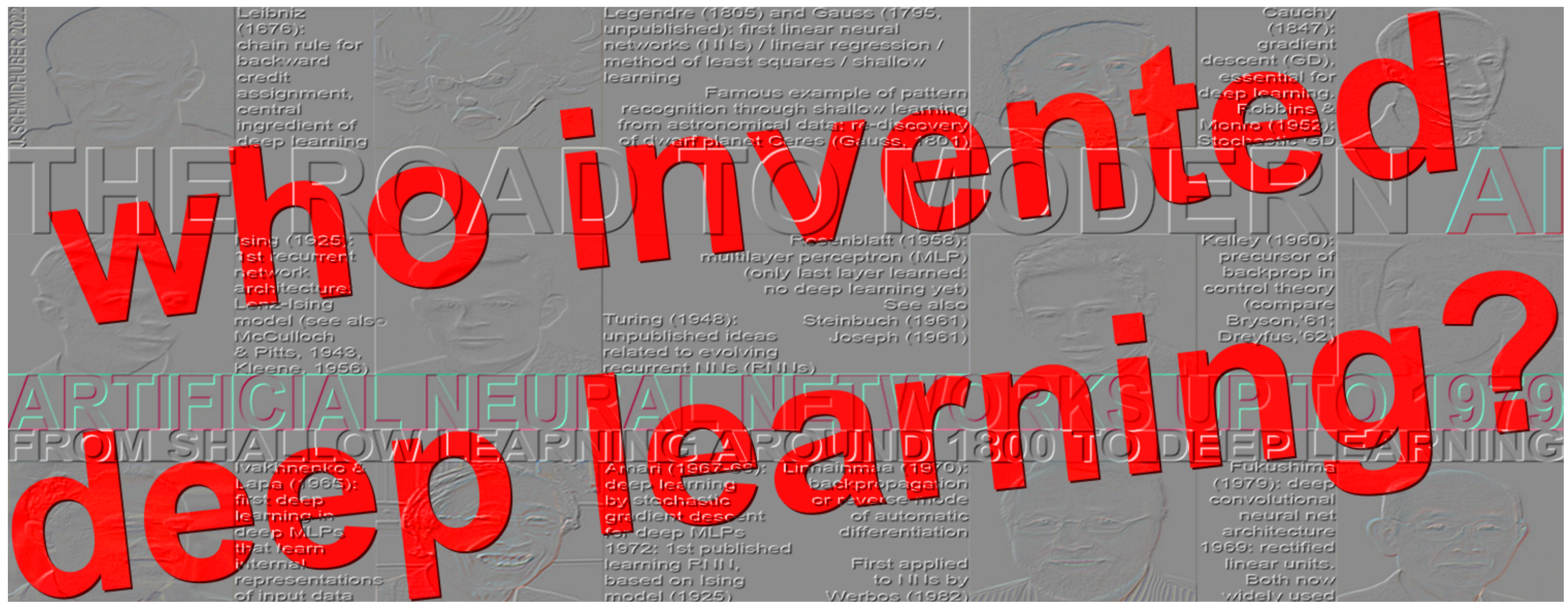

His pioneering work was repeatedly republished without attribution by researchers who went on to share a Turing award.[DLP][NOB][PLAG1-6][FAKE1-3] For example, the depth of Ivakhnenko's 1971 *layer-wise training*[DEEP2] was comparable to the depth of Hinton's and Bengio's 2006 *layer-wise training* published 35 years later[UN4][UN5] without comparison to the original work[NOB]—done when compute was millions of times more expensive. Similarly, LeCun et al.[LEC89] published NN pruning techniques without referring to Ivakhnenko's original work on pruning deep NNs. Even in their later "surveys" of deep learning,[DL3][DL3a] the awardees failed to mention the very origins of deep learning.[DLP][NOB] Ivakhnenko & Lapa also demonstrated that it is possible to learn appropriate weights for hidden units using only locally available information *without requiring a biologically implausible* backward pass.[BP4] 6 decades later, Hinton later attributed this achievement to himself.[NOB25a]

Why did deep learning emerge in the USSR in the early 1960s? Back then, the country was leading many important fields of science and technology, most notably in space: first satellite (1957), first man-made object on a heavenly body (1959), first man in space (1961), first woman in space (1962), first robot landing on a heavenly body (1965), first robot on another planet (1970). The USSR also detonated the world's biggest bomb ever (1961), and was home of many leading mathematicians, with sufficient funding for blue skies math research whose enormous significance would emerge only several decades later. See also: who invented deep learning?[DL25]

---

# 1967-68: Deep Learning by Stochastic Gradient Descent

---

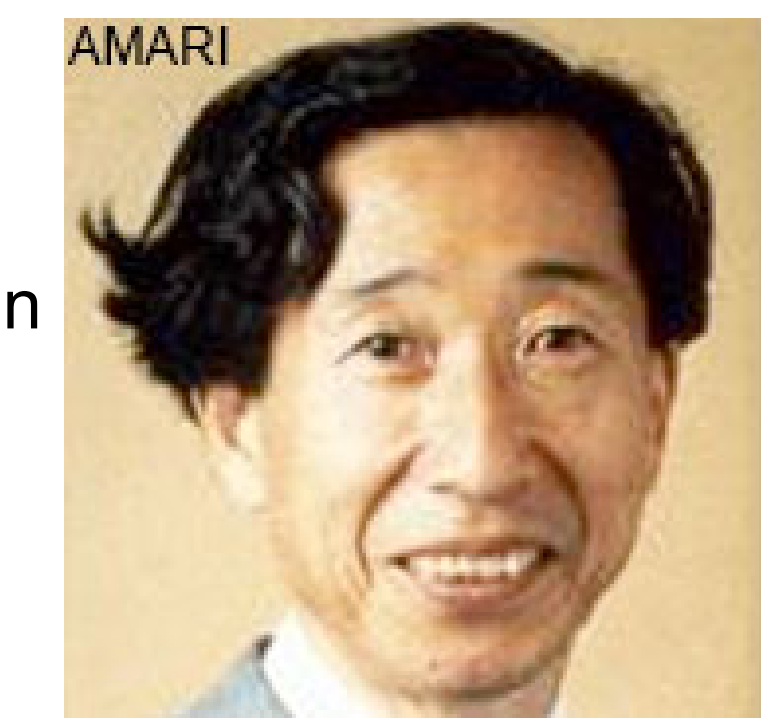

Ivakhnenko and Lapa (1965, see above) trained their deep networks layer by layer. In 1967, however, Shun-Ichi Amari suggested to train MLPs with many layers in non-incremental end-to-end fashion from scratch by stochastic gradient descent (SGD),[GD1] a method proposed in 1951 by Robbins & Monro.[STO51-52]

Amari's implementation[GD2,GD2a] (with his student Saito) learned *internal representations* in a five layer MLP with two modifiable layers, which was trained to classify non-linearily separable pattern classes. Back then compute was billions of times more expensive than today.

Note that Amari's method is general enough for *Reinforcement Learning* without a teacher. See Sec. 17.

See also Iakov Zalmanovich Tsypkin's even earlier work on gradient descent-based on-line learning for non-linear systems,[GDa-b] as well as the extensive work of his pupil Alexander Galushkin since 1970.[GAL07]

Remarkably, as mentioned above, Amari also published learning RNNs in 1972.[AMH1][NOB]

---

# 1970: Backpropagation. 1982: For NNs. 1960: Precursor.

---

In 1970, Seppo Linnainmaa was the first to publish what's now known as backpropagation, the famous algorithm for credit assignment in networks of differentiable nodes,[BP1,4,5] also known as "reverse mode of automatic differentiation." It is now the foundation of widely used NN software packages such as PyTorch and Google's Tensorflow.

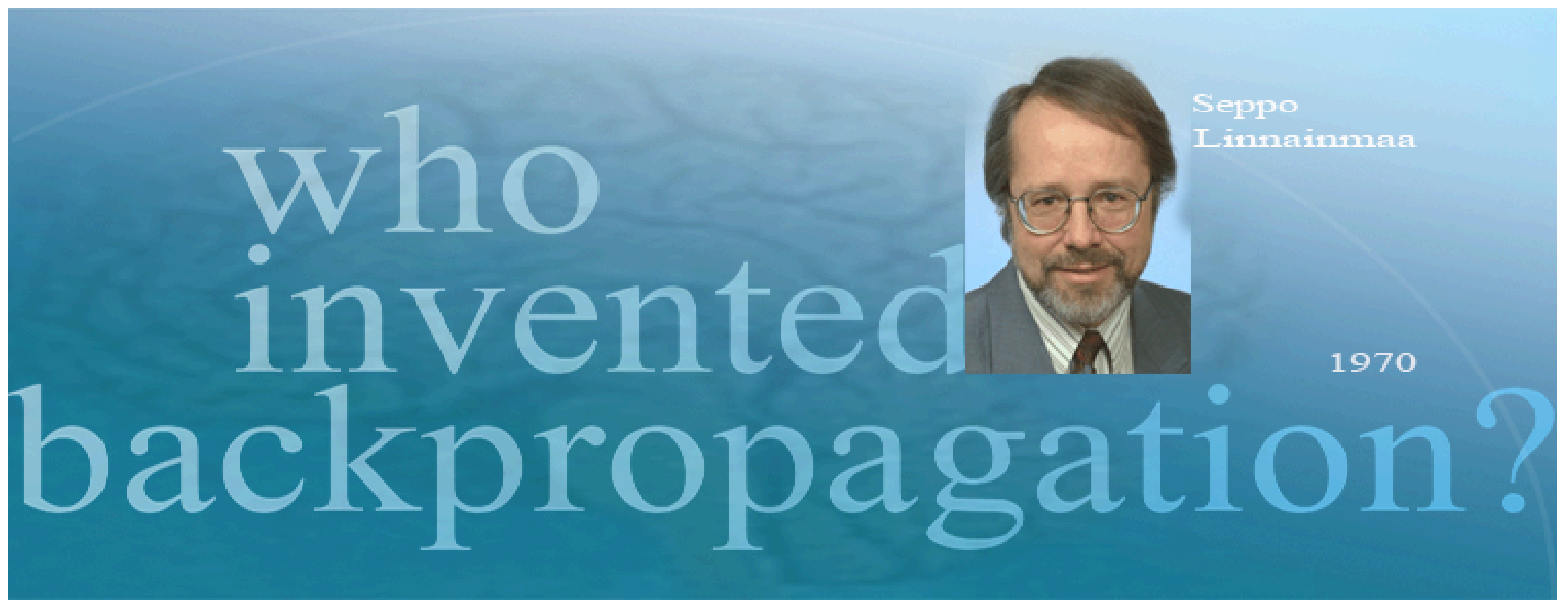

The first NN-specific application of efficient BP as above was described by Werbos in 1982[BP2] (but not yet in his 1974 thesis, as is sometimes claimed).

In 1960, Henry J. Kelley already had a precursor of backpropagation in the field of control theory;[BPA] see also later work of the early 1960s by Stuart Dreyfus and Arthur E. Bryson.[BPB][BPC][R7] Unlike Linnainmaa's general method,[BP1] the systems of the 1960s[BPA-C] backpropagated derivative information through standard Jacobian matrix calculations from one "stage" to the previous one, neither addressing direct links across several stages nor potential additional efficiency gains due to network sparsity.

KELLEY

Backpropagation is essentially an efficient way of implementing Leibniz's chain rule[LEI07-10] (1676) (see above) for deep networks (there are also many inefficient ways of doing this—see Footnote 3). Cauchy's gradient descent[GD'] uses this to incrementally weaken certain NN connections and strengthen others in the course of many trials, such that the NN behaves more and more like some teacher, which could be a human, or another NN,[UN-UN2] or something else.

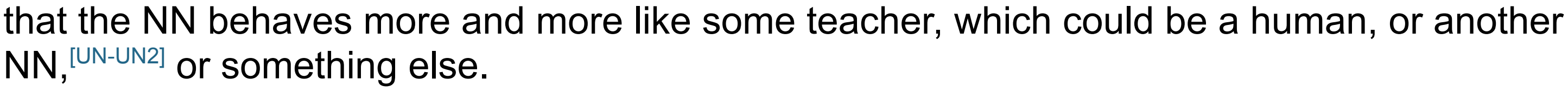

By 1985, compute had become about 1,000 times cheaper than in 1970, and the first desktop computers had just become accessible in wealthier academic labs. An experimental analysis of the known method[BP1-2][DLP][NOB] by David E. Rumelhart et al. then demonstrated that backpropagation can yield useful internal representations in hidden layers of NNs.[RUM] At least for supervised learning, backpropagation is generally more efficient than Amari's above-mentioned deep learning through the more general SGD method (1967), which learned useful internal representations in NNs about 2 decades earlier.[GD1-2a][DLP][NOB]

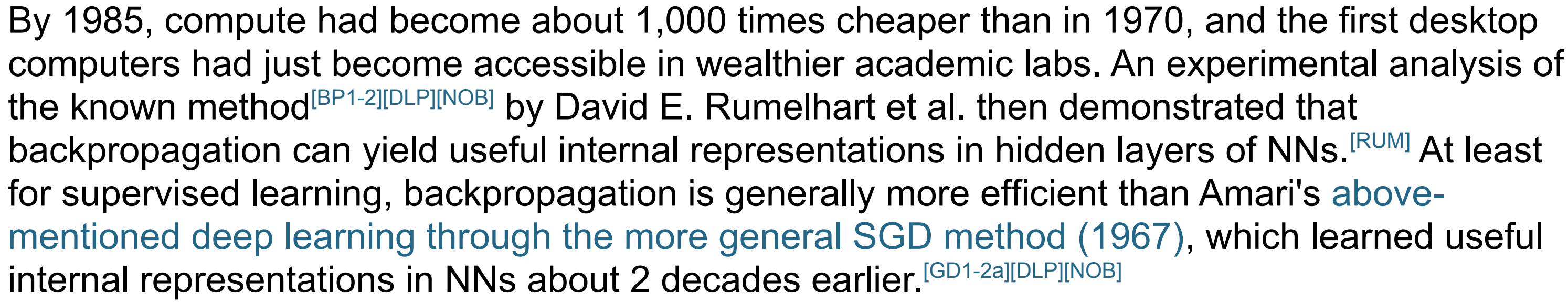

It took 4 decades until the backpropagation method of 1970[BP1-2] got widely accepted as a training method for *deep* NNs. Before 2010, many thought that the training of NNs with many layers requires unsupervised pre-training, a methodology introduced by Schmidhuber in 1991[UN][UN0-3] (see below), and later championed by others (2006).[UN4] In fact, it was claimed[VID1] that "nobody in their right mind would ever suggest" to apply plain backpropagation to deep

NNs. However, in 2010, the team with Schmidhuber's outstanding Romanian postdoc Dan Ciresan[MLP1-3] showed that deep FNNs can be trained on GPU by plain backpropagation and do not at all require unsupervised pre-training for important applications.[MLP3]

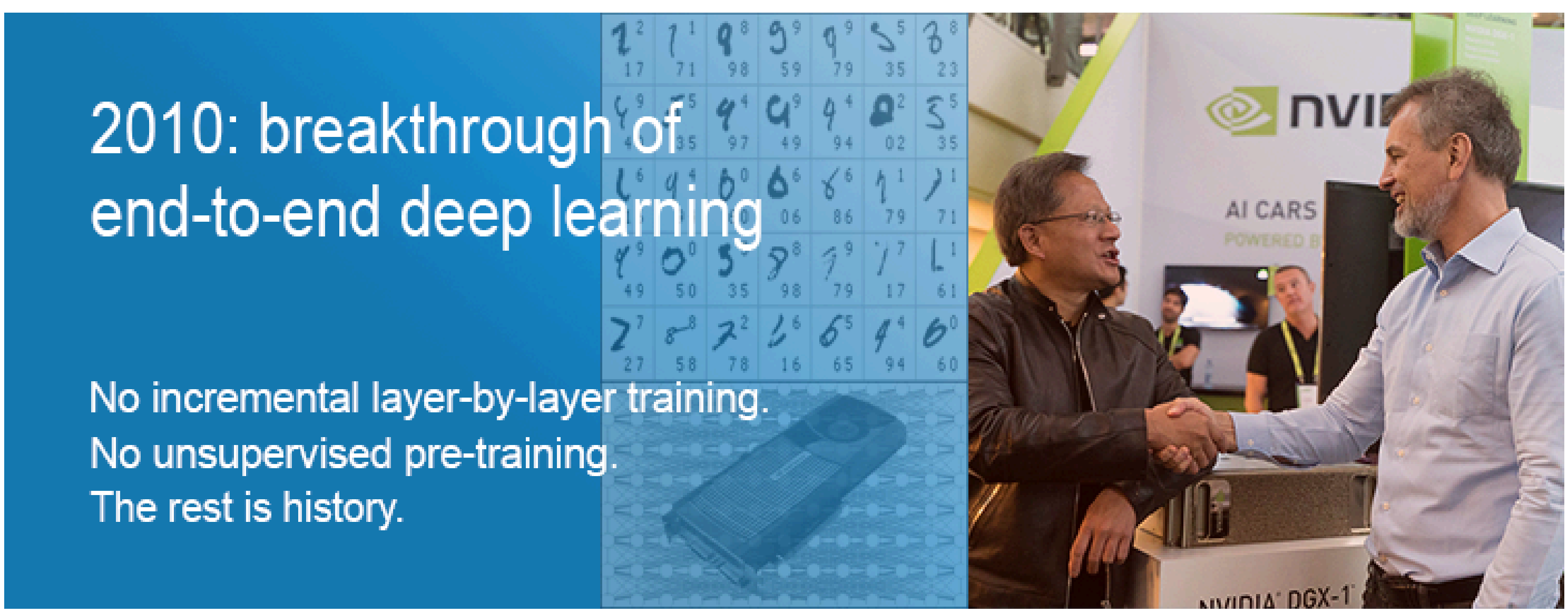

Their system set a new performance record[MLP1] on the back then famous and widely used image recognition benchmark called MNIST. This was achieved by greatly accelerating *deep* FNNs on highly parallel graphics processing units called GPUs (as first done for *shallow* NNs with few layers by Jung & Oh in 2004[GPUNN]). A reviewer called this a "wake-up call to the machine learning community." Today, everybody in the field is pursuing this approach.

---

*Footnote 5.* Unfortunately, several authors who republished backpropagation in the 1980s did not cite the prior art—not even in later surveys.[T22][DLP][NOB] In fact, as mentioned in the introduction, there is a broader, frequently repeated, misleading "history of deep learning"[S20] which ignores most of the pioneering work mentioned in the previous sections.[DLP][NOB][DLC] This "alternative history" essentially goes like this: *"In 1969, Minsky & Papert[M69] showed that shallow NNs without hidden layers are very limited and the field was abandoned until a new generation of neural network researchers took a fresh look at the problem in the 1980s."*[S20] However, the 1969 book[M69] addressed a "problem" of Gauss & Legendre's shallow learning (circa 1800)[NN25][DL1-2] that had already been solved 4 years prior by Ivakhnenko & Lapa's popular deep learning method,[DEEP1-2][DL2] and then also by Amari's SGD for MLPs.[GD1-2] Minsky neither cited this work nor corrected his book later.[HIN][DLP][NOB] And even recent papers promulgate this revisionist narrative of deep learning, apparently to glorify later contributions of their authors (such as the Boltzmann machine[BM][SK75][G63][DLP][NOB]) without relating them to the original work,[DLC][S20][DLP][NOB] although the true history is well-known. Deep learning research was alive and kicking in the 1960s-70s, especially outside of the Anglosphere.[DEEP1-2][GD1-3][CN79][DL1-2][DLP][NOB] Blatant misattribution and unintentional[PLAG1][CONN21] or intentional[FAKE2] plagiarism are still tainting the entire field of deep learning.[DLP][NOB] Scientific journals "need to make clearer and firmer commitments to self-correction,"[SV20] as is already the standard in other scientific fields.

# 1979: First Deep Convolutional NN (1969: ReLUs)

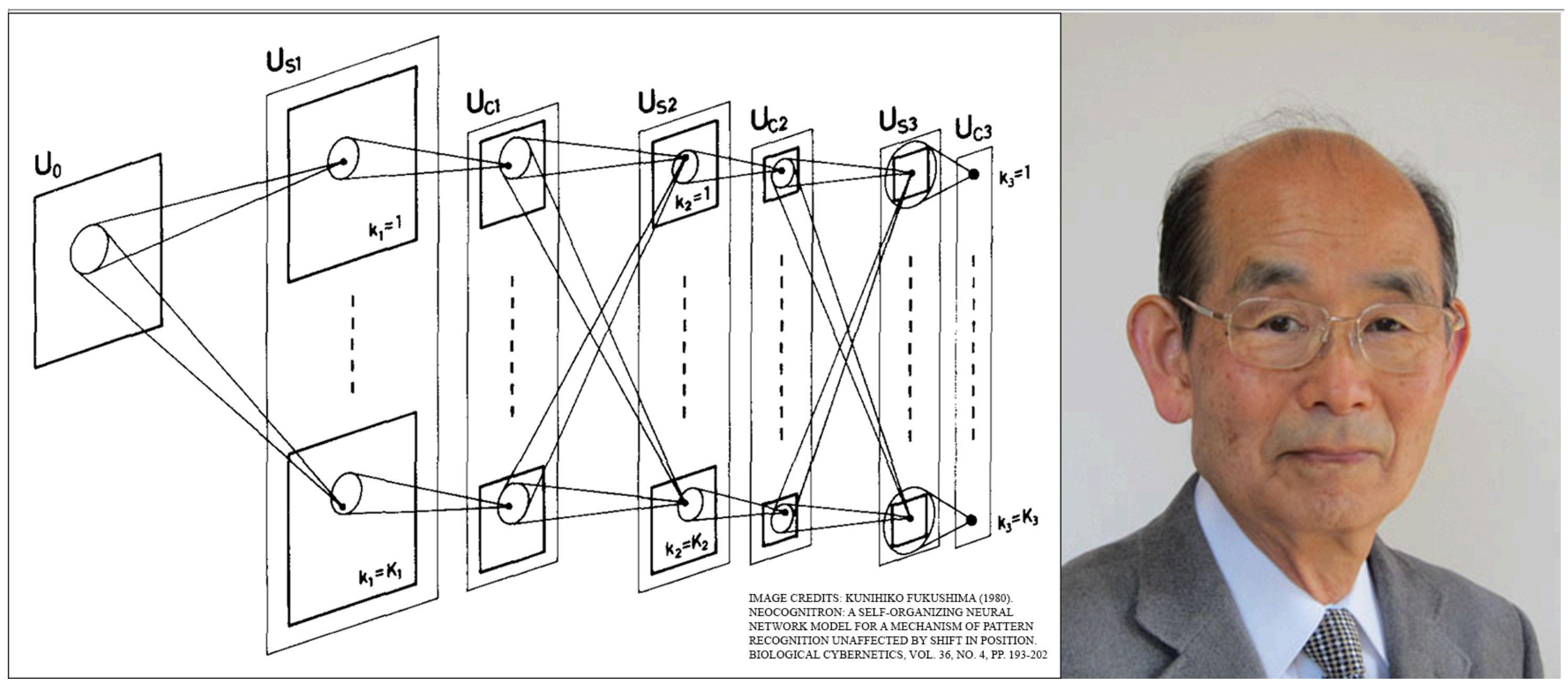

Computer Vision was revolutionized in the 2010s by a particular feedforward NN called the convolutional NN (CNN).[CNN79-25] The basic CNN architecture with alternating convolutional and downsampling layers is due to Kunihiko Fukushima (1979). He called it Neocognitron.[CN79] It was inspired by neurophysiological findings of Hubel and Wiesel,[HUW59-68] and trained by *unsupervised* learning rules. In 1986, Fukushima published a video on a CNN that recognizes handwritten digits[CN86] (video tweet).

In 1987, Alex Waibel (a German researcher working in Japan) trained *supervised* weight sharing NNs with 1-dimensional convolutions (TDNNs) by Linnainmaa's 1970 backpropagation algorithm[BP1-5] to recognise speech.[CN87][CN89c] A similar proposal by Homma et al.[CN87b] introduced the *"convolution"* terminology to NNs.

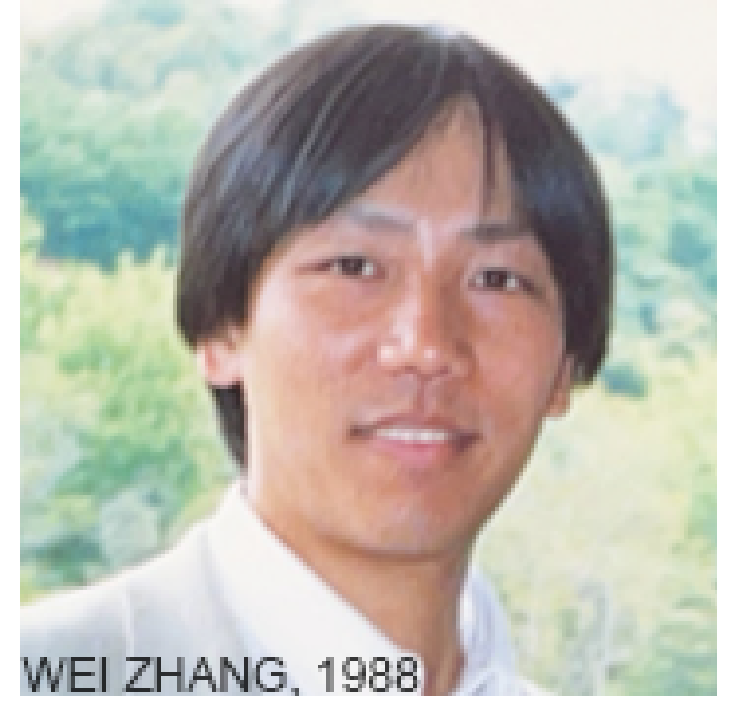

In 1988, Wei Zhang (a Chinese researcher working in Japan) and colleagues had *the first "modern" 2-dimensional CNN trained by backpropagation*, and applied it to character recognition.[CN88] Compute was about 10 million times more expensive than today.

Most of the above was published in Japan 1979-1988.[CN79][CN87][CN88] Why Japan? Let's look back at the 1980s. Back then, Japan was the envy of the world. Before the 1990 crash,[95-25] the Tokyo stock market was the world's largest, and the world's 6 most valuable public companies were all Japanese. So were the world's richest business men. According to the real estate market, tiny Japan was 4 times more valuable than the much bigger US. The central square mile of Tokyo had the value of California. Japan had far more robots than any other country,[95-25] and by far the most expensive AI project: the 5th Generation Project. Interestingly, this project had little to do with neural networks; it was mostly about logic programming and expert systems. So Fukushima and colleagues were outsiders back then. However, Japan already had a strong tradition and role models in NN research—see, e.g., Amari's above-mentioned pioneering work of the 1960s (Sec. 7) and 1970s (Sec. 4)—and at least there was sufficient funding in Japan, even for such unpopular types of blue skies research. Today, the rest of the world can be thankful for that.[CN25]

Zhang et al. also had the first journal submission on "modern" backpropagation-trained CNNs (with applications to character recognition).[CN89] In the early 1990s, Zhang et al. published several additional important CNN papers.[CN91-CN94] Yann LeCun et al. at Bell Labs had the second journal submission on backpropagation-trained CNNs for character recognition (zip codes),[CN89b][DLP] following the work of Zhang et al.[CN88][CN89] See also Hampshire & Waibel (1989).[CN89d]

In 1990-93, Fukushima's downsampling based on spatial averaging[CN79] was replaced by *max-pooling* for 1-D convolutional NNs (Yamaguchi et al.)[CN90] and for 2-D CNNs (Weng et al.).[CN93] See also: Who invented convolutional neural networks?[CN25]

Many additional CNN papers were published in the 1990s and early 2000s.[e.g., CN98-CN10] For example, Baldi and Chauvin (1993) had the first application of CNNs with backpropagation to biomedical/biometric images.[BA93]

Already in 1969, Fukushima described **rectified linear units** or ReLUs[CN69] which are now extensively used in CNNs. See also Householder's work[HOU41] (1941) on nerve-fiber networks with piecewise linear activation functions.

CNNs became more popular in the ML community much later in 2011 when Schmidhuber's team greatly sped up the training of deep CNNs (Dan Ciresan et al., 2011).[GPUCNN1,3,5][DAN][CN25b] The fast GPU-based[GPUNN][GPUCNN5] CNN of 2011[GPUCNN1] known as DanNet[DAN,DAN1][R6] was a practical breakthrough, much deeper and faster than earlier GPU-accelerated CNNs of 2006.[GPUCNN] In 2011, DanNet became the first pure deep CNN to win computer vision contests.[GPUCNN2-3,5] Admittedly, however, this was mostly about engineering & scaling up the basic insights from the previous millennium, profiting from much faster hardware.

| Competition[GPUCNN5] | Date/Deadline | Image size | Improvement | Winner |
|---|---|---|---|---|
| ICDAR 2011 Chinese handwriting | May 15, 2011 | variable | 3.8% / 28.9% | DanNet[GPUCNN1-3] |
| IJCNN 2011 traffic signs | **Aug 06, 2011** | variable | **68.0%** (superhuman) | **DanNet**[DAN,DAN1][R6] |
| ISBI 2012 image segmentation | Mar 01, 2012 | 512x512 | 26.1% | DanNet[GPUCNN3a] |
| ICPR 2012 medical imaging | Sep 10, 2012 | 2048x2048x3 | 8.9% | DanNet[GPUCNN8] |
| ImageNet 2012 | Sep 30, 2012 | 256x256x3 | 41.4% | AlexNet[GPUCNN4] |
| MICCAI 2013 Grand Challenge | Sep 08, 2013 | 2048x2048x3 | 26.5% | DanNet[GPUCNN8] |
| ImageNet 2014 | Aug 18, 2014 | 256x256x3 | | VGG Net[GPUCNN9] |
| ImageNet 2015 | Sep 30, 2015 | 256x256x3 | 15.8% | ResNet,[HW2] like a Highway Net[HW1] with open gates |

For a while, DanNet enjoyed a monopoly. From 2011 to 2012 it won every contest it entered, winning four of them in a row (15 May 2011, 6 Aug 2011, 1 Mar 2012, 10 Sep 2012).[GPUCNN5] In particular, at IJCNN 2011 in Silicon Valley, DanNet crushed the competition, performing three times better than the closest competitor (by LeCun's team), and twice as good as humans.[DAN1] DanNet was also the first deep CNN to win: a Chinese handwriting contest (ICDAR 2011), an image segmentation contest (ISBI, May 2012), a contest on object detection in large images (ICPR, 10 Sept 2012), and— at the same time—a medical imaging contest on cancer detection.[GPUCNN8] In 2010, we introduced DanNet to Arcelor Mittal, the world's largest steel producer, and were able to greatly improve steel defect detection.[ST] To the best of my knowledge, this was the first deep learning breakthrough in heavy industry. Most computer vision researchers knew about DanNet in late 2011 / early 2012.[CN25b] In July 2012, the CVPR paper on DanNet[GPUCNN3] hit the computer vision community. 5 months later, the similar GPU-accelerated AlexNet won the ImageNet[IM09] 2012 contest.[GPUCNN4-5][R6][NOB] Masci et al.'s CNN image scanners were 1000 times faster than previous methods.[SCAN] This attracted tremendous interest from the healthcare industry. Today IBM, Siemens, Google and many startups are pursuing this approach. The VGG network (ImageNet 2014 winner)[GPUCNN9] and other highly cited CNNs[RCNN1-3] further extended the DanNet of 2011.[MIR](Sec. 19)[MOST]

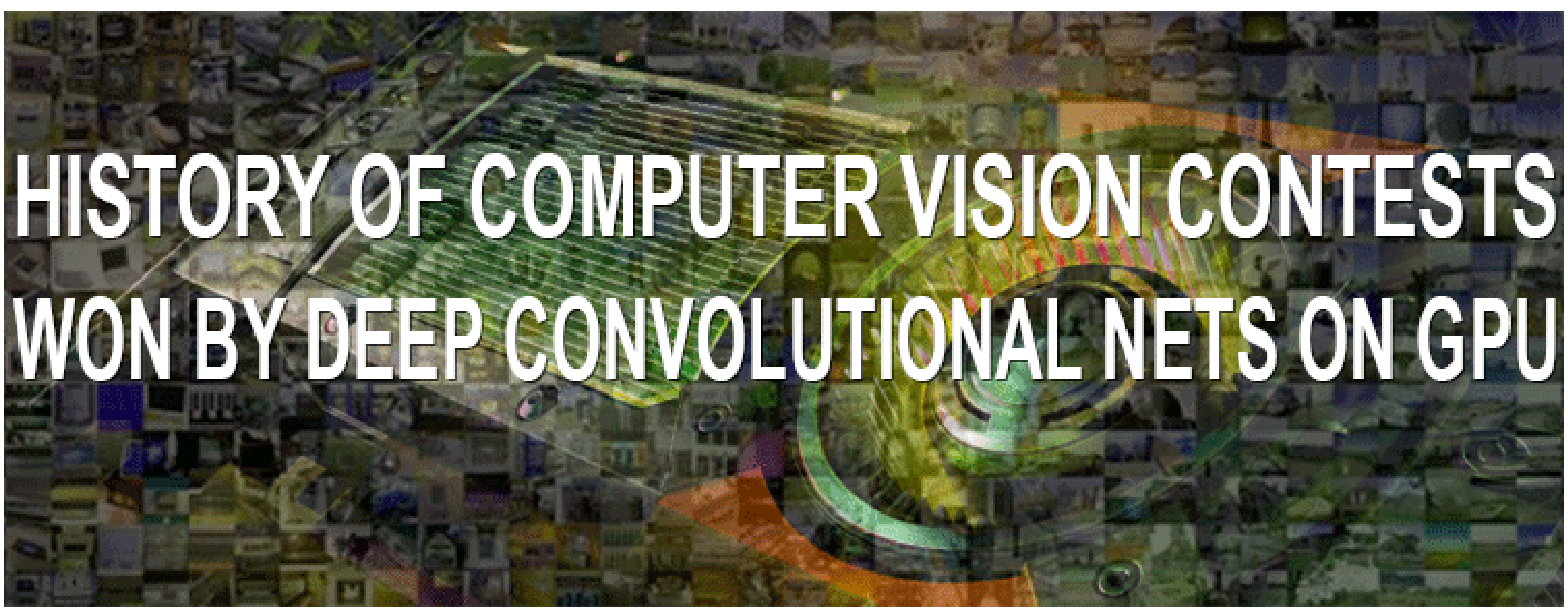

ResNet, the ImageNet 2015 winner[HW2] (Dec 2015) and currently the most cited NN,[MOST] is an open-gated variant of the earlier Highway Net (May 2015).[HW1-3][R5] The Highway Net (see below) is actually the feedforward net version of the vanilla LSTM (see below).[LSTM2] It was the first working, really deep feedforward NN with hundreds of layers (previous NNs had at most a few tens of layers).

---

## 1980s-90s: Graph NNs / Stochastic Delta Rule (Dropout) /...

---

Apart from the developments mentioned above, the last two decades of the past millennium brought additional important NN-related insights.

NNs with rapidly changing "fast weights" were introduced by v.d. Malsburg (1981) and others.[FAST,a,b] Deep learning architectures that can manipulate structured data such as graphs[T22] were proposed in 1987 by Pollack[PO87-90] and extended/improved by Sperduti, Goller, and Küchler in the early 1990s.[SP93-97][GOL][KU][T22] See also The graph NN-like, Transformer-like Fast Weight Programmers of 1991[ULTRA][FWP0-1][FWP6][FWP] which learn to continually rewrite mappings from inputs to outputs (addressed below), and the work of Baldi and colleagues.[BA96-03] Today, graph NNs are used in numerous applications.

Werbos,[BP2][BPTT1] Williams,[BPTT2][CUB0-2] and others[ROB87][BPTT3][DL1] analyzed ways of implementing gradient descent[GD'][STO51-52][GDa-b][GD1-2a] in RNNs. Kohonen's self-organising maps became popular.[KOH82-89]

As mentioned in Sec. 6, in the 1960s, it was demonstrated that it is possible to learn appropriate weights for hidden neurons without requiring a biologically implausible backward pass.[DEEP1-2][NOB] The 80s and 90s saw various additional proposals of biologically more plausible deep learning algorithms that—unlike backpropagation—are local in space and time.[BB2][NAN1-4][NHE][CC90][HEL] See overviews[MIR](Sec. 15, Sec. 17) and recent renewed interest in such methods.[NAN5][FWPMETA6][HIN22][DLP][NOB][NOB25a]

In 1990, Hanson introduced the Stochastic Delta Rule, a stochastic way of training NNs by backpropagation. Decades later, a variant of this became popular under the moniker "dropout."[Drop1-4][GPUCNN4][NOB]

Many additional papers on NNs (including RNNs) were published in the 1980s and 90s—see the numerous references in the 2015 survey.[DL1] Here, however, we mostly limit ourselves to the—in hindsight—most essential ones, given the present (ephemeral?) perspective of 2025.

Interestingly, in the Anglosphere, the 1980s were considered a key decade for NN research and *"connectionism,"* although the most foundational NN concepts were developed earlier *outside* of the Anglosphere, as shown in previous sections.

---

# Feb 1990: Generative Adversarial Networks / Curiosity

---

Generative Adversarial Networks (GANs) have become very popular.[MOST] For example, many modern *deepfakes*[GAN19b] were created by GANs, which also have many beneficial applications, e.g., in healthcare.

The first NNs that were both generative and adversarial were published in 1990-1991[MIR] by Juergen Schmidhuber at a time when compute was about 10 million times more expensive than today (2025).[GAN90][GAN91] How did they work? There are two NNs that fight each other. A so-called *controller NN* with adaptive stochastic Gaussian units (a *generative model*) generates output data. This output is fed into a *predictor NN* (called a World Model in 1990[GAN90][GAN91][PLAN]) which learns by gradient descent to predict the effects of the outputs. However, in a minimax game, the generator NN *maximizes* the error/loss *minimized* by the predictor NN.

So the controller is motivated to create through its outputs experiments/situations that *surprise* the predictor. As the predictor improves, these situations become *boring*. This in turn incentivizes the controller to invent new outputs (or experiments) with still less predictable outcomes, and so forth. This was called Artificial Curiosity.[GAN91][GAN10][AC] The world model can also be used for continual online action planning.[GAN90][PLAN2-3][PLAN] See Sec. 17.

Artificial Curiosity wasn't the first adversarial machine learning setting, but earlier works[e.g.,][S59][H90] were *very different*—they neither involved self-supervised NNs where one NN sees the output of another generative NN and tries to predict its consequences, nor were about modeling data, nor used gradient descent. (Generative models themselves are much older, e.g., Hidden Markov Models.[MM1-3])

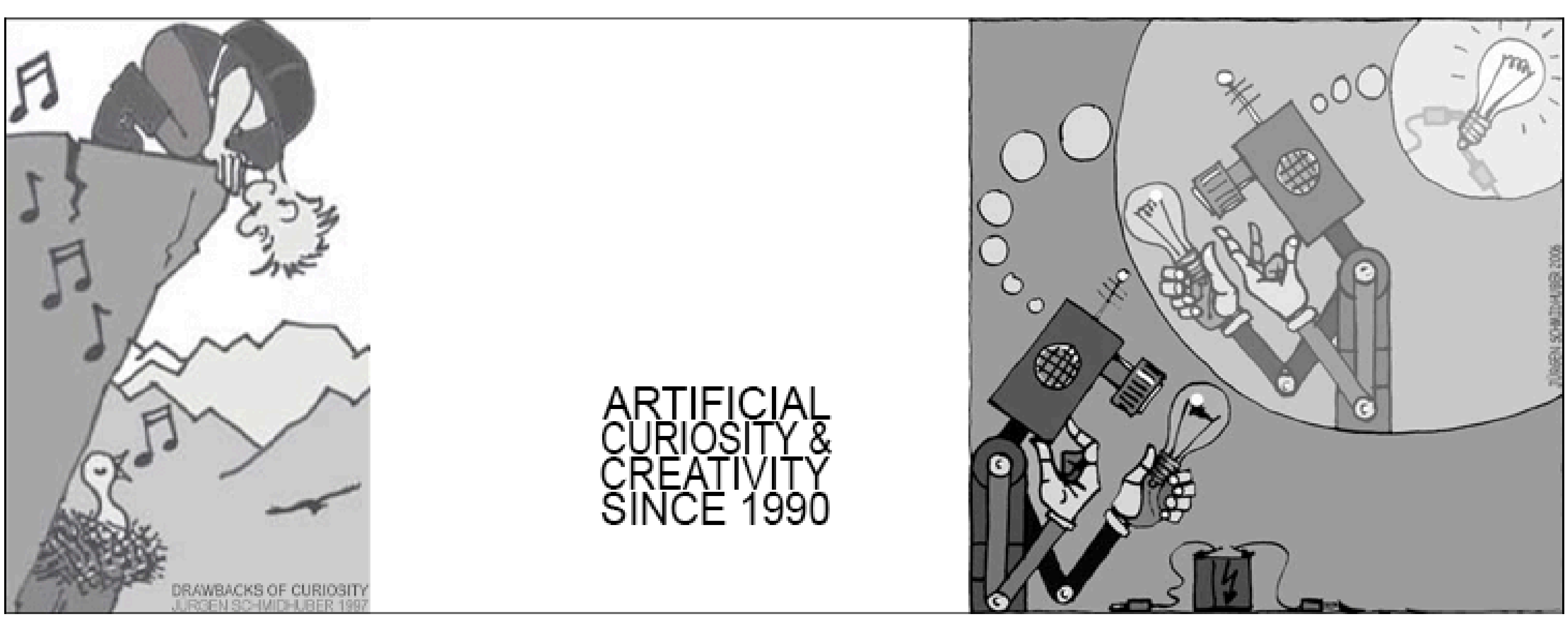

See Section *"Implementing Dynamic Curiosity and Boredom"* of the 1990 technical report[GAN90] and the 1991 peer-reviewed conference paper.[GAN91] It mentions preliminary experiments where (in absence of external reward) the predictor minimizes a linear function of what the generator maximizes. So these old papers essentially describe what would become known as a GAN almost a quarter of a century later, in 2014,[GAN14] when compute was about 100,000 times cheaper than in 1990. In 2014, the 1990 neural *predictor or world model*[GAN90][GAN91] was called a *discriminator*,[GAN14] predicting *binary* effects of possible outputs of the generator (such as *real* vs *fake*).[GAN20] The applications to image generation[GAN10b][GAN14] were novel. The 1990 GAN was more general than the 2014 GAN: it wasn't limited to *single* output actions in 1-step trials, but permitted long *sequences* of actions. More sophisticated generative adversarial systems for artificial curiosity & creativity were published in 1997, predicting *abstract internal representations* instead of raw data.[AC97][AC99][AC02][LEC]

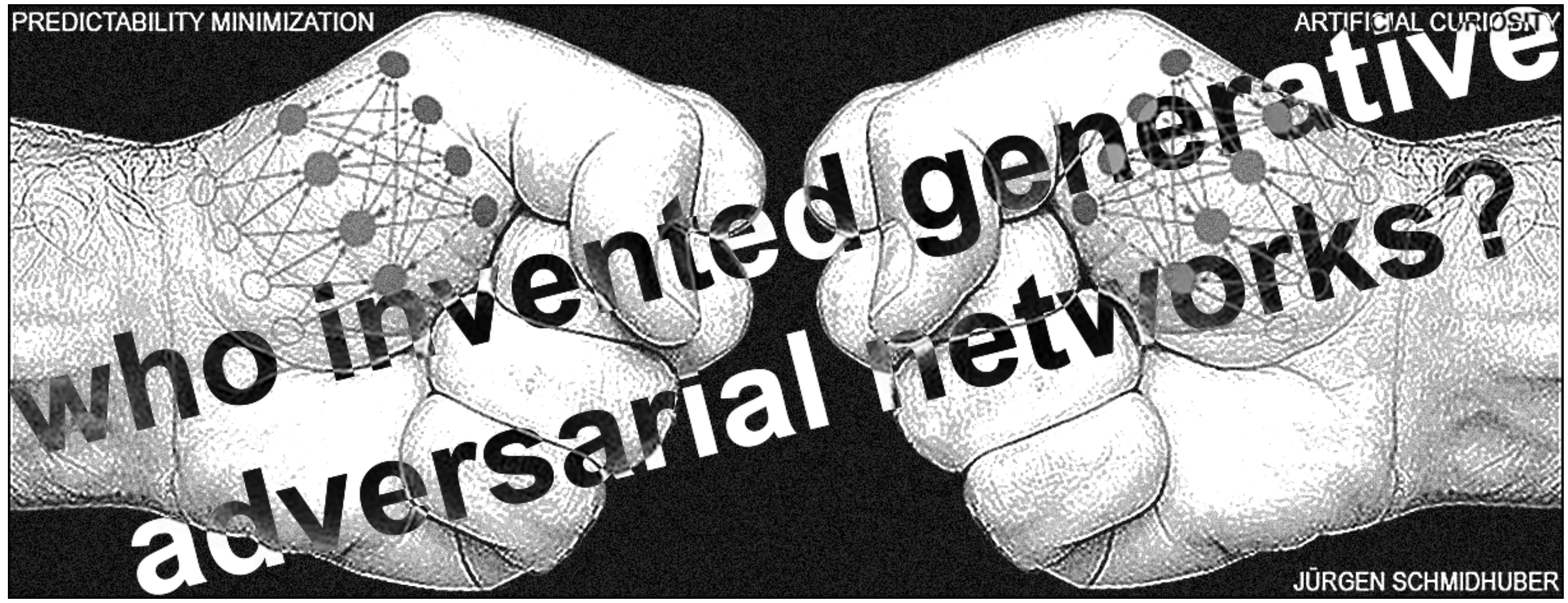

The 1990 principle has been widely used for exploration in Reinforcement Learning[SIN5][OUD13][PAT17][BUR18] and for synthesis of realistic images,[GAN1,2] although the latter domain was recently taken over by Rombach et al.'s *Latent Diffusion*, another method published in Munich,[DIF1] building on Jarzynski's earlier work in physics from the previous millennium[DIF2] and more recent papers.[DIF3-5]

In 1991, Schmidhuber published yet another ML method based on two adversarial NNs called Predictability Minimization for creating disentangled representations of partially redundant data, applied to images in 1996.[PM0-2][GAN20][R2][MIR](Sec. 7) As of 2025, a 2014 paper[GAN14] on generative adversarial neural networks (GANs) is the most cited research paper of Turing awardee Dr. Bengio,[DLP][MOST] although it failed to cite the original 1990 work on generative and adversarial neural networks.[GAN90][GAN91][R2][GAN20][DLP] A paper on who invented generative adversarial networks[GAN25] summarizes this GAN priority dispute.

---

# April 1990: NNs Generate Subgoals / Work on Command

---

Most NNs of recent centuries were dedicated to simple pattern recognition, not to high-level reasoning, which is now considered a remaining grand challenge.[LEC] The early 1990s, however, saw first exceptions: NNs that learn to *decompose* complex spatio-temporal observation sequences into compact but meaningful *chunks*[UN0-3] (see further below), and NN-based planners of hierarchical action sequences for *compositional learning*,[HRL0] as discussed next. This work injected concepts of traditional *"symbolic"* hierarchical AI[NS59][FU77] into end-to-end differentiable *"subsymbolic"* NNs.

In 1990, Schmidhuber's NNs learned to generate hierarchical action plans with end-to-end differentiable NN-based subgoal generators for Hierarchical Reinforcement Learning (HRL).[HRL0] Soon afterwards, this was also done with recurrent NNs that learn to generate sequences of subgoals.[HRL1-2][PHD][MIR](Sec. 10) An RL machine gets extra *command inputs* of the form *(start, goal)*. An evaluator NN learns to predict the current rewards/costs of going from *start* to *goal*. An (R)NN-based subgoal generator also sees *(start, goal)*, and uses (copies of) the evaluator NN to learn by gradient descent a sequence of cost-minimising intermediate subgoals. The RL machine tries to use such subgoal sequences to achieve final goals. The system is learning action plans at multiple levels of abstraction and multiple time scales and solves (at least in principle) what has been called an "open problem" in 2022.[LEC]

Compare other NNs that have "worked on command" since April 1990, in particular, for learning selective attention,[ATT0-3] artificial curiosity and self-invented problems,[PP][PPa,1,2][AC] upside-down reinforcement learning[UDRL1-2] and its generalizations.[GGP]

---

# March 1991: Unnormalized Linear Transformers

---

The T in ChatGPT[GPT3] stands for a famous NN called *Transformer*. While reading long texts, it learns by backpropagation[BP4] to create sequences of context-dependent KEY and VALUE patterns focusing its internal attention on relevant short-lived memories queried by incoming data, to better predict the next word in context-dependent fashion, given the text so far.

According to Google Scholar (2025), the scientific article that is cited most frequently each year appears to be a 2017 paper on Transformers. Who invented this?

In 1991, Schmidhuber published the original tech report on what's now called the unnormalized linear Transformer (ULTRA).[FWP0][ULTRA] KEY/VALUE was called FROM/TO. ULTRA uses *outer product rules* to associate its self-invented KEYs/VALUEs through fast weights,[FAST][FWP] and applies the resulting context-dependent attention mappings to incoming queries. ULTRA's computational costs scale *linearly* in input size, that is, for 1,000 times more text we need 1,000 times more compute, which is acceptable. Like modern *quadratic* Transformers (see below), the 1991 ULTRA is highly parallelizable. It was a by-product of more general research on NNs that learn to program fast weight changes of other NNs,[FWP,FWP0-9,FWPMETA1-10] back then called *fast weight controllers*[FWP0] or *fast weight programmers (FWPs)*.[FWP] ULTRA was presented as an alternative to recurrent NNs.[FWP0] The 1991 experiments were similar to today's: predict some effect, given a sequence of inputs.[FWP0]

In 1993, a recurrent ULTRA extension[FWP2] introduced the terminology of learning *"internal spotlights of attention."*

In 2014, end-to-end sequence-to-sequence models[S2Sa,b,c,d] became popular for *Natural Language Processing*. They were *not* based on the 1991 unnormalized linear Transformer[ULTRA] above, but on the Long Short-Term Memory (LSTM) recurrent NN from the same lab.[LSTM0-13] In 2014, this approach was combined with an attention mechanism[ATT14] that isn't *linearized* like the 1991-93 attention[FWP0-2] but includes a *nonlinear* softmax operation. The first *Large Language Models* (LLMs) were based on such LSTM-attention systems. See additional work on attention from 2016-17.[ATT16a-17b]

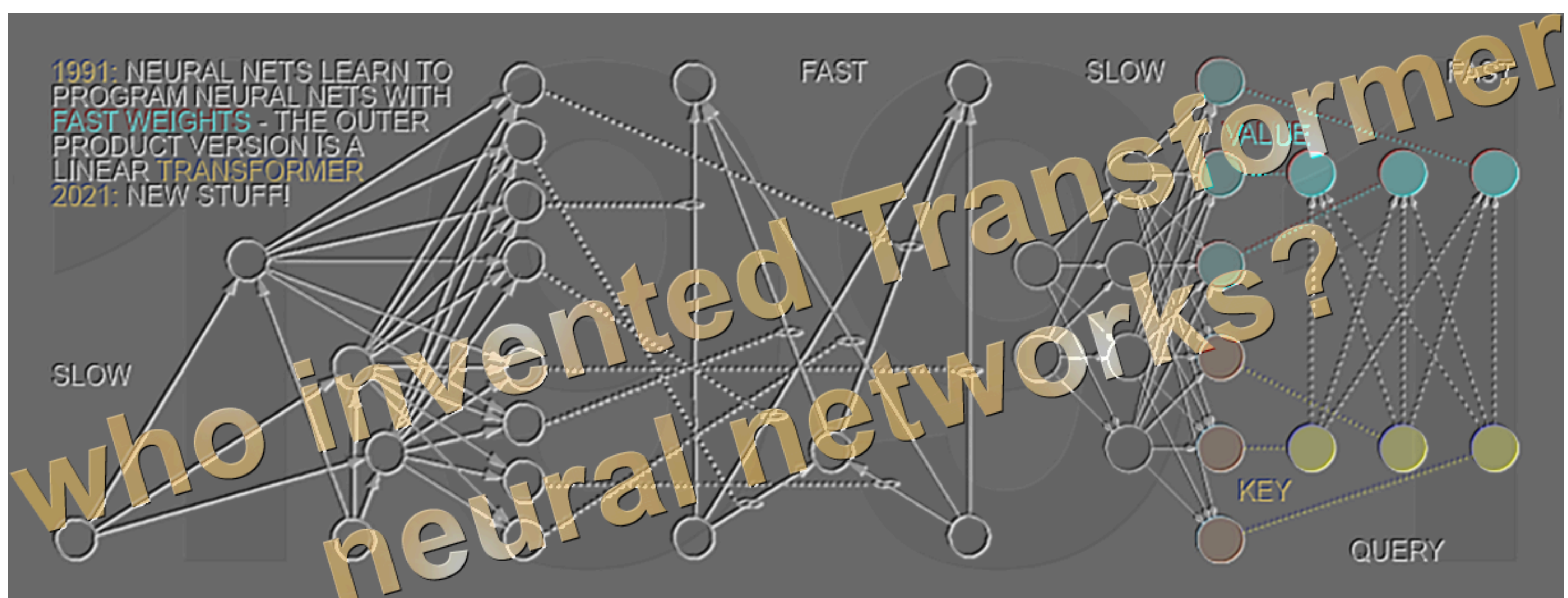

In 2017, the "modern" *quadratic* Transformer (*"attention is all you need"*) was published by Vaswani et al.,[TR1] scaling *quadratically* in input size, that is, for 1,000 times more text one needs 1,000,000 times more compute. Note that in 1991,[ULTRA] no journal would have accepted an NN that scales quadratically, but by 2017, compute was cheap enough to apply the quadratic Transformer (a kind of fast weight programmer[FWP]) to large amounts of data on massively parallel computers. The quadratic Transformer combines the 1991 additive outer product fast weight principle[FWP0-2] and *softmax* (see 2014 above): *attention* (query, KEY, VALUE) ~ *softmax* (query KEY) VALUE.

In 2020, a paper[TR5] used the terminology *"linear Transformer"* for a more efficient Transformer variant that scales *linearly*, leveraging *linearized attention.*[TR5a] In 2021, it was pointed out that the unnormalised *linear Transformer*[TR5-6] is actually *mathematically equivalent* to a 1991 *fast weight controller*[FWP0][ULTRA] published when compute was a million times more expensive than in 2021. See also 2022 tweet for ULTRA's 30-year anniversary, and 2024 tweet. See also: who invented transformer neural networks?[TR25]

In 2021-25, work on extensions of ULTRAs and other FWPs (such as the DeltaNet[FWP6]) has become mainstream research, aiming to develop sequence models that are both efficient and powerful.[TR6,TR6a][LT23-25][FWP23-25b]

Today's Transformers heavily use unsupervised pre-training for deep NNs[UN0-3] (the P in ChatGPT—see next section), another deep learning methodology published in the Annus Mirabilis of 1990-1991.[MIR][MOST]

The 1991 fast weight programmers also led to meta-learning self-referential NNs that can run their own weight change algorithm or learning algorithm on themselves, and improve it, and improve the way they improve it, and so on. This work since 1992[FWPMETA1-10][HO1] extended Schmidhuber's 1987 diploma thesis,[META1] which introduced algorithms not just for learning but also for meta-learning or learning to learn,[META][META10][METARL2-10] to learn better learning algorithms through experience. This became popular in the 2010s[DEC] when computers were a million times faster.

---

# April 1991: Deep Learning by Self-Supervised Pre-Training

---

Today's most powerful NNs tend to be very deep, that is, they have many layers of neurons or many subsequent computational stages.[MIR] Before the 1990s, however, gradient-based training did not work well for deep NNs, only for shallow ones[DL1-2] (but see a 1989 paper[MOZ]). This *Deep Learning Problem* was most obvious for *recurrent* NNs. Like the human brain, but unlike the more limited *feedforward* NNs (FNNs), RNNs have feedback connections. This makes RNNs powerful, general purpose, parallel-sequential computers that can process input sequences of arbitrary length (think of speech data or videos). RNNs can in principle implement any program that can run on your laptop or any other computer in existence. If we want to build an *Artificial General Intelligence* (AGI), then its underlying computational substrate must be something more like an RNN than an FNN as FNNs are fundamentally insufficient; RNNs and similar systems are to FNNs as general computers are to pocket calculators. In particular, unlike FNNs, RNNs can in principle deal with problems of arbitrary depth.[DL1] Before the 1990s, however, RNNs failed to learn deep problems in practice.[MIR](Sec. 0)

To overcome this drawback through RNN-based *"general deep learning,"* Schmidhuber built a self-supervised RNN hierarchy that learns representations at multiple levels of abstraction and multiple self-organizing time scales:[LEC] the *Neural Sequence Chunker*[UN0] or Neural History Compressor.[UN1] Each RNN tries to solve the *pretext task* of predicting its next input, sending only unexpected inputs (and therefore also targets) to the next RNN above. The resulting compressed sequence representations greatly facilitate downstream *supervised* deep learning such as sequence classification. Such *unsupervised or self-supervised pre-training for deep NNs* is now widely used—see the P in ChatGPT.

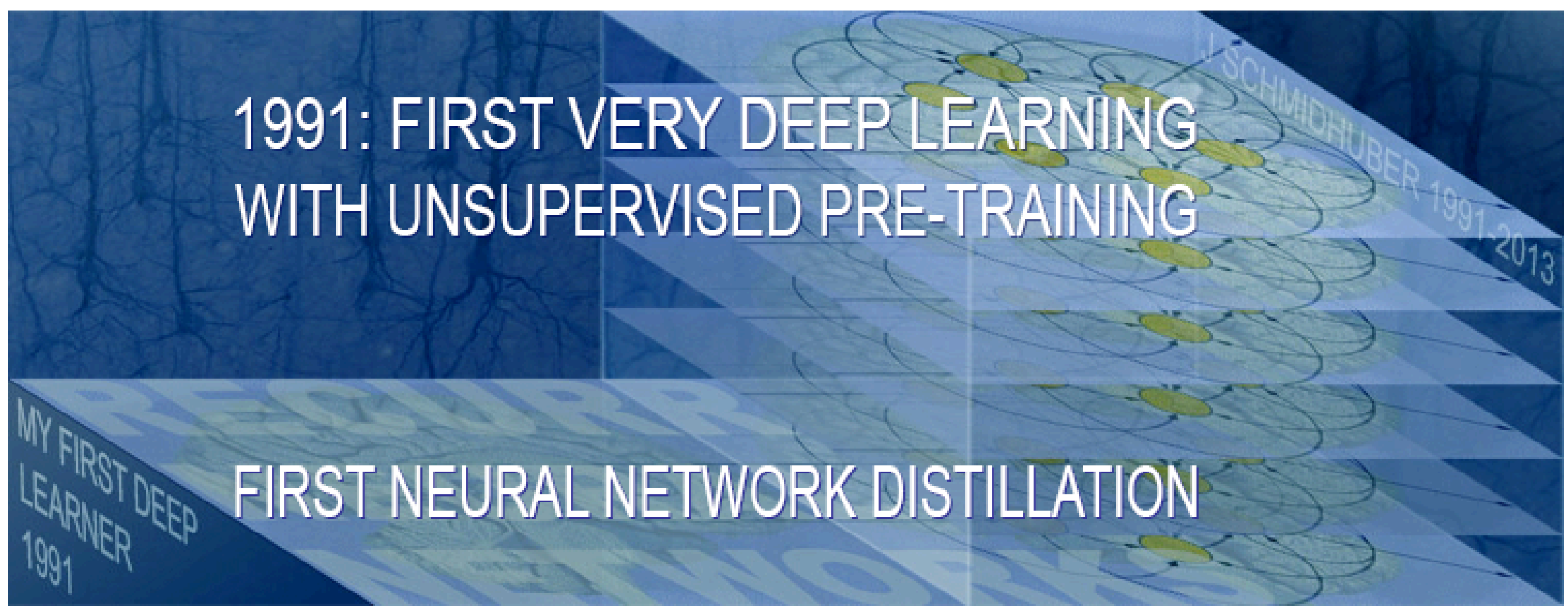

Although computers back then were about a million times slower per dollar than today, by 1993, the Neural History Compressor above was able to solve previously unsolvable "very deep learning" tasks of depth > 1000[UN2] (requiring more than 1,000 subsequent computational stages—the more such stages, the deeper the learning). In 1993, a *continuous* version of the Neural History Compressor was published.[UN3] (See also recent work on unsupervised NN-based abstraction.[OBJ1-5])

More than a decade after this work,[UN1] a similar unsupervised method for more limited *feedforward* NNs (FNNs) was published, facilitating supervised learning by unsupervised pre-training of stacks of FNNs called *Deep Belief Networks* (DBNs).[UN4] The 2006 justification was essentially the one used in the early 1990s for the RNN stack: each higher level tries to reduce the description length (or negative log probability) of the data representation in the level below.[HIN][T22][MIR]

## April 1991: Distilling one NN into another NN

In January 2025, the NN-based *DeepSeek "Sputnik"*[DS1] shocked the commercial AI scene and wiped out a trillion USD from the stock market. DeepSeek[DS1] and many other Large Language Models use NN distillation to transfer knowledge from one NN to another. Who invented this?

NN distillation was published by Schmidhuber in 1991.[UN0-3][UN][MIR][DLP] See Section 4 of the paper on the "conscious" chunker and a "subconscious" automatiser,[UN0][UN1] which introduced a general principle for transferring the knowledge of one NN to another. Suppose a teacher NN has learned to predict (conditional expectations of) data, given other data (like the neural history compressor above). Its knowledge can be compressed into a student NN, by training[BP1-5,A-C] the student NN to imitate the behavior of the teacher NN (while also re-training the student NN on previously learned skills such that it does not forget them).

In 1991, this was called *"collapsing"* or *"compressing"* one NN into another. Today, this is widely used, and also referred to as *"distilling"*[DIST2][HIN][DLP] or *"cloning"* the behavior of a teacher NN into that of a student NN. It even works when the NNs are recurrent and operate on different time scales.[UN0][UN1] See also related work[DIST3-4] and: who invented knowledge distillation with artificial neural networks?[DIST25]

DeepSeek also used elements of Schmidhuber's 2015 reinforcement learning (RL) prompt engineer[PLAN4] and its 2018 refinement[PLAN5] which collapses the 2015 RL machine and its world model[PLAN4] into a single net through the NN distillation of 1991: a distilled *chain of thought* system. See the popular tweet of 31 Jan 2025.

Today, unsupervised pre-training is heavily used by Transformers[TR1][TR25] (the T in ChatGPT) for natural language processing and other domains. Remarkably, Transformers with linearized self-attention were also first published[ULTRA] in the Annus Mirabilis of 1990-1991,[MIR][MOST] together with unsupervised/self-supervised pre-training for deep learning.[UN0-3] See the previous section.

---

## June 1991: Fundamental Problem: Vanishing Gradients

---

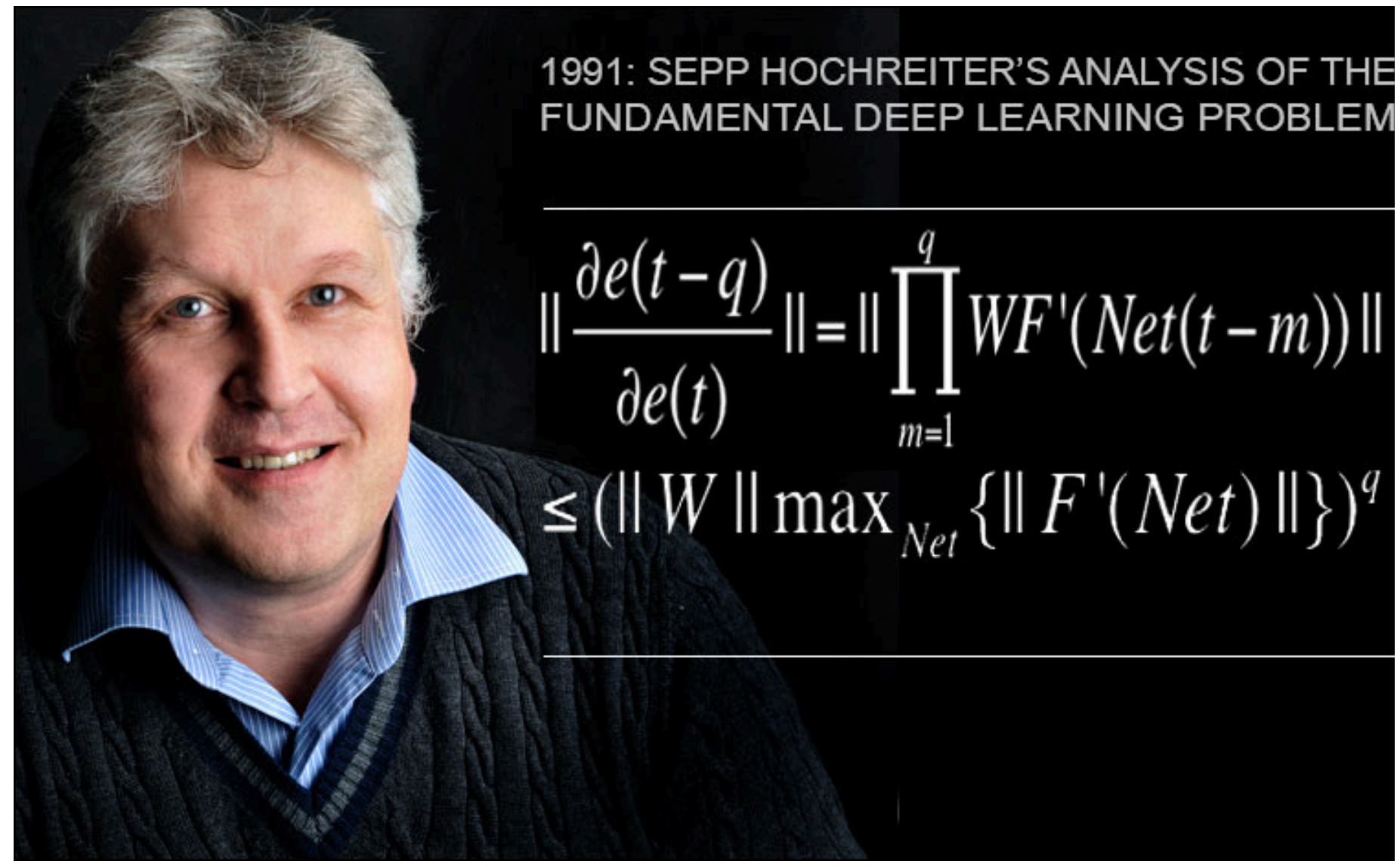

Deep learning is hard because of the Fundamental Deep Learning Problem identified and analyzed in 1991 by Sepp Hochreiter in a diploma thesis (June 1991)[VAN1] supervised by Schmidhuber, at a time when compute was about 10 million times more expensive than today (2025). First he implemented the Neural History Compressor above but then did much more: he showed that deep NNs suffer from the now famous problem of vanishing or exploding gradients: in typical deep or recurrent networks, back-propagated error signals either shrink rapidly, or grow out of bounds. In both cases, learning fails.[VAN1][DLP][DLH]

To solve the vanishing gradient problem, Hochreiter mathematically derived from first principles what's now called a recurrent residual connection:[HW25] a neural unit with the *identity activation function* has a connection to itself, and the weight of this connection is 1.0. This simple setup ensures *constant error flow* in deep gradient-based RNNs: errors can be backpropagated [BP1-4][BPTT1-2] through such units for millions of steps without vanishing or exploding,[VAN1] since according to the 1676 chain rule[LEI07-21b][L84] by Leibniz (see Sec. 2), the relevant multiplicative first derivatives (and their weights) are always 1.0. This insight led to basic principles of what's now called LSTM and Highway Nets / ResNets (see below).

# June 1991: Roots of LSTM / Highway Nets / ResNets

The Long Short-Term Memory (LSTM) recurrent neural network[LSTM1-6] overcomes the Fundamental Deep Learning Problem identified by Hochreiter in his above-mentioned 1991 diploma thesis,[VAN1] one of the most important documents in the history of machine learning. It also provided essential insights for overcoming the problem, through basic principles (such as *constant error flow* along residual connections[HW25]) of what Schmidhuber called LSTM in a tech report of 1995.[LSTM0] The main peer-reviewed publication of 1997[LSTM1][25y97] is now the most cited AI paper of the 20th century.[MOST] The LSTM core units with residual connections (weight 1.0) were called *constant error carrousels* (CECs).[LSTM1]

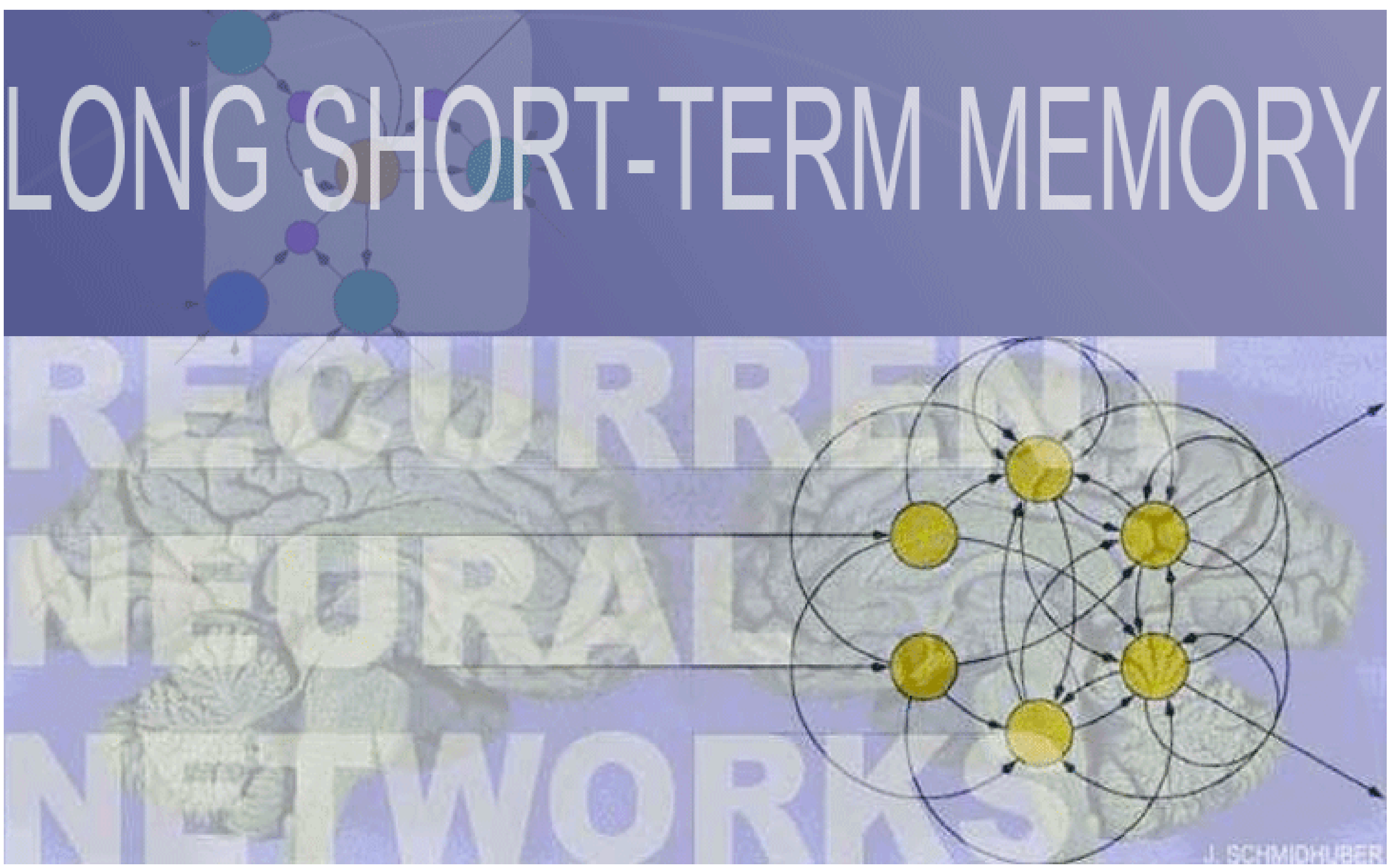

LSTM and its training procedures were further improved on Schmidhuber's Swiss LSTM grants at IDSIA through the work of his later students Felix Gers, Alex Graves, and others. A milestone was the *"vanilla LSTM architecture"* with forget gate[LSTM2]—the LSTM variant of 1999-2000 that everybody is using today, e.g., in Google's Tensorflow. It features gated recurrent residual connections whose gates are initially open (1.0) such that they start out as *plain* residual connections.[HW25] Graves was lead author of the first successful application of LSTM to speech (2004).[LSTM10] 2005 saw the first publication of LSTM with full backpropagation through time and of bi-directional LSTM[LSTM3] (now widely used): this led from *recurrent* to *feedforward* residual NNs.[HW25]

Another milestone of 2006 was the training method *"Connectionist Temporal Classification"* or CTC[CTC] for simultaneous alignment and recognition of sequences. CTC-trained LSTM was successfully applied to speech in 2007[LSTM4] (also with hierarchical LSTM stacks[LSTM14]). This led to the first superior end-to-end neural speech recognition. It was very different from hybrid methods since the late 1980s which combined NNs and traditional approaches such as Hidden Markov Models (HMMs).[BW][BRI][BOU][HYB12][T22][DLP] In 2009, through the efforts of Alex, LSTM trained by CTC became the first RNN to win international competitions, namely, three ICDAR 2009 Connected Handwriting Competitions (French, Farsi, Arabic). This attracted enormous interest from industry. LSTM was soon used for everything that involves sequential data such as speech[LSTM10-11][LSTM4][DL1] and videos. In 2015, the CTC-LSTM combination dramatically improved Google's speech recognition on the Android smartphones.[GSR15] Many other companies adopted this.[DL4] Google's on-device speech recognition of 2019 (on the phone, not on the server) is still based on LSTM.

## 1995-: Neural Probabilistic Language Model

The first superior end-to-end neural machine translation was also based on LSTM. In 1995, Schmidhuber & Heil already had a *neural probabilistic text model* for excellent text compression[SNT] whose basic concepts were reused in 2003[NPM][T22]—see also Pollack's earlier work on embeddings of words and other structures[PO87][PO90] as well as Nakamura and Shikano's 1989 word category prediction model.[NPMa] In 2001, LSTM learned languages unlearnable by traditional models such as HMMs,[LSTM13] i.e., a neural "subsymbolic" model suddenly excelled at learning "symbolic" tasks. Compute still had to get 1000 times cheaper, but by 2016, Google Translate[GT16]—whose whitepaper[WU] mentions LSTM over 50 times—was based on two connected LSTMs,[S2Sa,b,c,d] one for incoming texts, and one for outgoing translations—much better than what existed before.[DL4] By 2017, LSTM also powered Facebook's machine translation (over 30 billion translations per week—the most popular youtube video needed years to achieve only 10 billion clicks),[FB17][DL4] Apple's Quicktype on roughly 1 billion iPhones,[DL4] the voice of Amazon's Alexa,[DL4] Google's image caption generation[DL4] & automatic email answering[DL4] etc. Business Week called LSTM "arguably the most commercial AI achievement."[AV1] By 2016, more than a quarter of the awesome computational power for inference in Google's datacenters was used for LSTM (and 5% for another popular Deep Learning technique called CNNs—see above).[JOU17] And of course, LSTM is also massively used in healthcare and medical diagnosis—a simple Google Scholar search turns up innumerable medical articles that have "LSTM" in their title.[DEC] The first Large Language Models (LLMs) were based on LSTM as well.

## May 2015: Very Deep Feedforward NNs

While supervised LSTM RNNs had become very deep in the 1990s through residual connections, backpropagation-based FNNs had remained rather shallow until 2014: they had at most 20-30 layers or so, despite massive help through fast GPU-based hardware.[MLP1-3][DAN,DAN1][GPUCNN1-9]

Since depth is essential for deep learning, the principles of deep LSTM RNNs were transferred to deep FNNs. In May 2015, the resulting Highway Networks[HW1][HW1a] were the first working really deep gradient-based FNNs with hundreds of layers, over ten times deeper than previous FNNs. They worked because they adapted the 1999 LSTM principle of *gated* recurrent residual connections (gates initially open: 1.0)[LSTM2a][LSTM2] from RNNs to FNNs.[HW25] This work was conducted by Schmidhuber's PhD students Rupesh Kumar Srivastava and Klaus Greff. Its principles have become very popular.

As of 2025, the most cited scientific article of the 21st century is a paper on *deep residual learning* (Dec 2015)[HW1-HW25] with *residual NNs* containing *residual connections*:[MOST25,25b] the *ResNet*[HW2] (which won the ImageNet 2015 contest) is like a *Highway Net* variant whose gates are always open. In turn, the Highway Net is a *gated ResNet*. The gates of the *gated residual connections* of Highway Nets are initially open (1.0) such that the network starts out with *plain* residual connections (weight 1.0) which allow for very deep error propagation like in LSTM's unfolded CECs—this is what makes Highway NNs (and ResNets) so deep. The earlier Highway Nets perform roughly as well as their ResNet variants on ImageNet.[HW3][HW25]

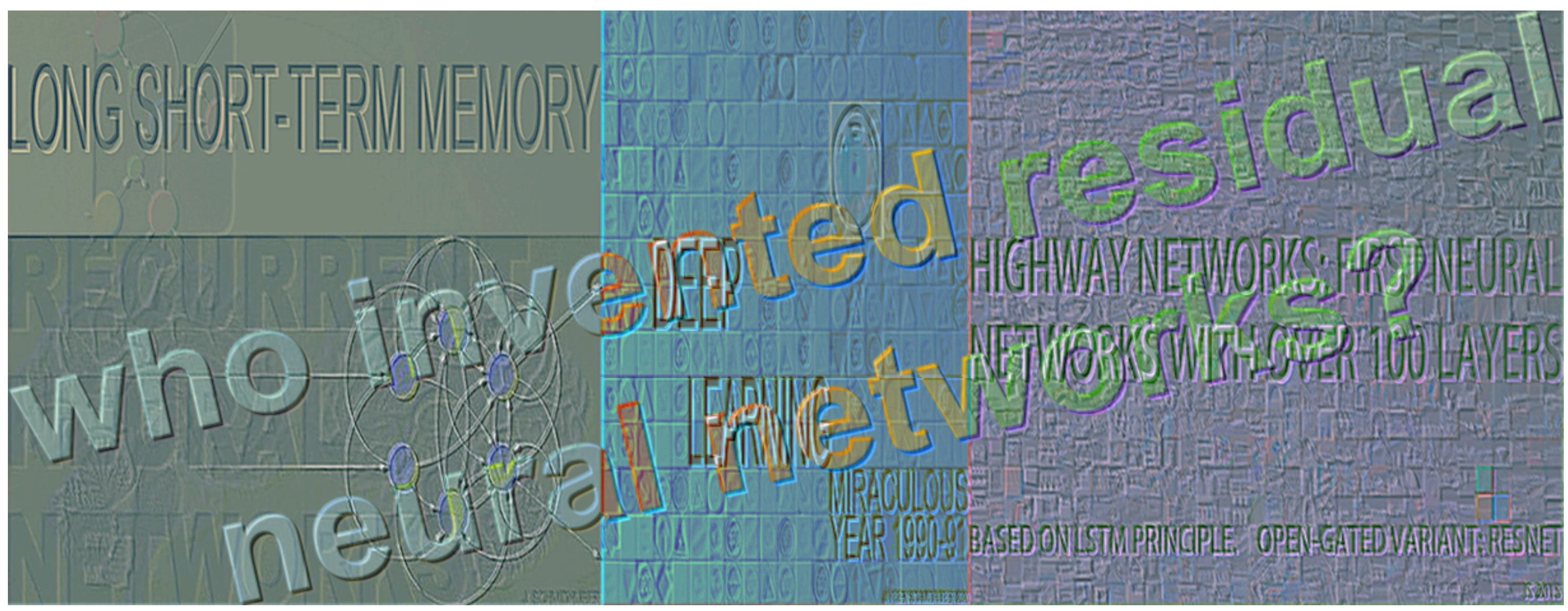

## The LSTM / Highway Net Principle is the Core of Modern Deep Learning

Deep learning is all about NN depth.[DL1] In the 1990s, the residual connections of LSTMs brought essentially *unlimited* depth to supervised *recurrent* NNs; in the 2010s, the gated (initially open) residual connections of LSTM-inspired Highway Nets brought it to *feedforward* NNs. LSTM has become the most cited NN of the 20th century; the Highway Net variant called ResNet the most cited NN of the 21st.[MOST] (Citations, however, are a highly questionable measure of true impact.[NAT1]) LSTM-like and Highway Net-like NNs can deal with very deep credit assignment paths (CAPs)[DL1] and are now massively used. Here is a compact recap of the timeline of their evolution taken from a separate report: who invented deep residual learning?[HW25]

★ 1991: Hochreiter's recurrent residual connections solve the vanishing gradient problem[VAN1]
★ 1997 LSTM: *plain* recurrent residual connections (weight 1.0)[LSTM0-1]
★ 1999 LSTM: *gated* recurrent residual connections (gates initially open: 1.0)[LSTM2a][LSTM2]
★ 2005: unfolding LSTM—from *recurrent* to *feedforward* residual NNs[LSTM3]
★ May 2015: deep Highway Net—gated *feedforward* residual connections (initially 1.0)[HW1]
★ Dec 2015: ResNet—like an open-gated Highway Net (or an unfolded 1997 LSTM)[HW2][HW25]

The basic LSTM principle of *constant error flow through residual connections* is central not only to deep RNNs but also to deep FNNs. And it all dates back to 1991.[MIR][HW25]

---

# 1980s-: NNs for Learning to Act Without a Teacher

---

The previous sections have mostly focused on deep learning for passive pattern recognition/classification. However, NNs are also relevant for Reinforcement Learning (RL),[KAE96][BER96][TD3][UNI][GM3][LSTMPG] the most general type of learning. General RL agents must discover, without the aid of a teacher, how to interact with a dynamic, initially unknown, partially observable environment in order to maximize their expected cumulative reward signals.[DL1] There may be arbitrary, a priori unknown delays between actions and perceivable consequences. The RL problem is as hard as any problem of computer science, since any task with a computable description can be formulated in the general RL framework.[UNI]

Certain RL problems can be addressed through non-neural techniques invented long before the 1980s: Monte Carlo (tree) search (MC, 1949),[MOC1-5] dynamic programming (DP, 1953),[BEL53] artificial evolution (1954),[EVO1-7]([TUR1],unpublished) alpha-beta-pruning (1959),[S59] control theory and system identification (1950s),[KAL59][GLA85] stochastic gradient descent (SGD, 1951),[STO51-52] and universal search techniques (1973).[AIT7]

Deep FNNs and RNNs, however, are useful tools for *improving* certain types of RL. In the 1980s, concepts of function approximation and NNs were combined with system identification,[WER87-89][MUN87][NGU89] DP and its online variant called Temporal Differences (TD),[TD1-3] artificial evolution,[EVONN1-3] and policy gradients.[GD1][PG1-3] Many additional references on this can be found in Sec. 6 of the 2015 survey.[DL1]

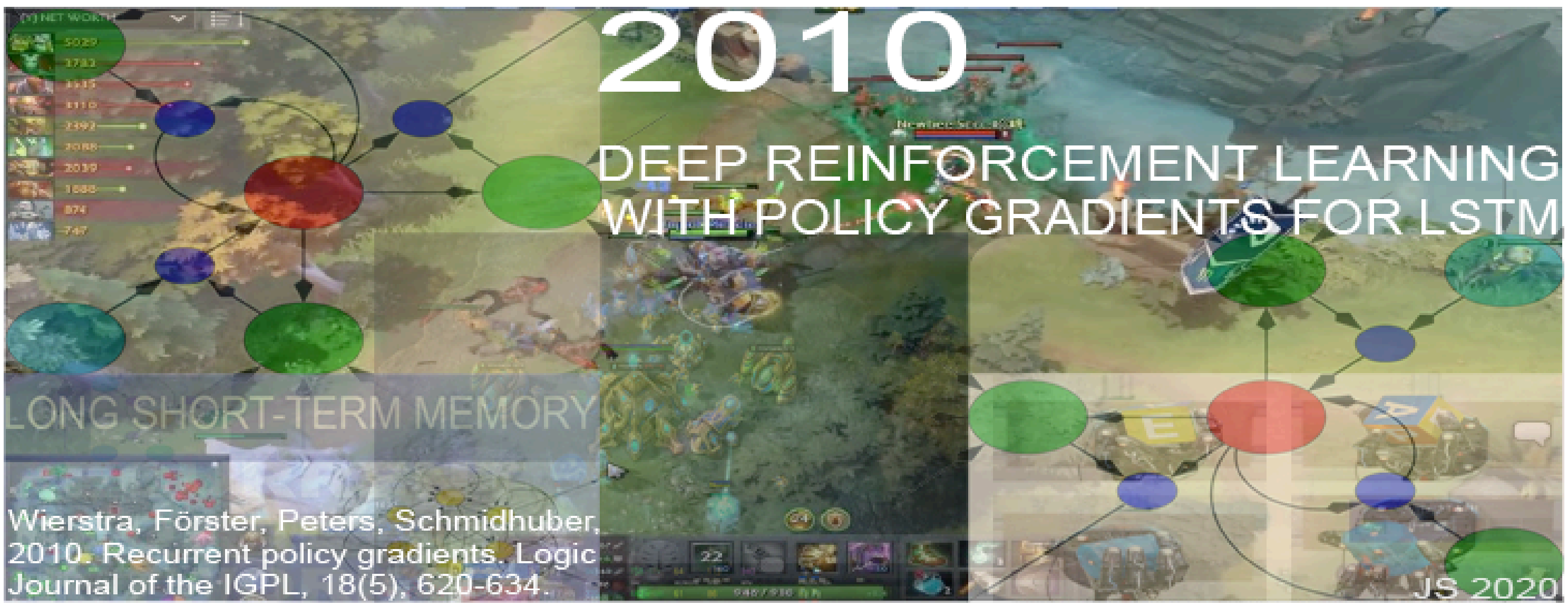

When there is a Markovian interface[PLAN3] to the environment such that the current input to the RL machine conveys all the information required to determine a next optimal action, RL with DP/TD/MC-based FNNs can be very successful, as shown in 1994[TD2] (master-level backgammon player) and the 2010s[DM1-2a] (superhuman players for Go, chess, and other games).

For more complex cases without Markovian interfaces, where the learning machine must consider not only the present input, but also the history of previous inputs, combinations of RL algorithms and LSTM[LSTM-RL][RPG] have become standard, in particular, LSTM trained by policy gradients (2007).[RPG07][RPG][LSTMPG]

For example, in 2018, a PG-trained LSTM was the core of OpenAI's famous Dactyl which learned to control a dextrous robot hand without a teacher.[OAI1][OAI1a] Similar for video games: in 2019, DeepMind (co-founded by a student from Schmidhuber's lab) famously beat a pro player in the game of Starcraft, which is theoretically harder than Chess or Go[DM2] in many ways, using Alphastar whose brain has a deep LSTM core trained by PG.[DM3] An RL LSTM (with 84% of the model's total parameter count) also was the core of the famous OpenAI Five which learned to defeat human experts in the Dota 2 video game (2018).[OAI2] Bill Gates called this a *"huge milestone in advancing artificial intelligence"*.[OAI2a][MIR](Sec. 4)[LSTMPG]

The future of RL will be about learning/composing/planning with compact spatio-temporal abstractions of complex input streams—about commonsense reasoning[MAR15] and *learning to think*.[PLAN4-5] How can NNs learn to represent percepts and action plans in a hierarchical manner, at multiple levels of abstraction, and multiple time scales?[LEC][DLP] Schmidhuber published first answers to these questions in 1990-91: self-supervised neural history compressors[UN][UN0-3] learn to represent percepts at multiple levels of abstraction and multiple time scales (see above), while end-to-end differentiable NN-based subgoal generators[HRL3][MIR](Sec. 10) learn hierarchical action plans through gradient descent (see above).

More sophisticated ways of learning to think in abstract ways were published in 1997[AC97][AC99][AC02] and 2015-18.[PLAN4-5] A few years later, in 2025, the *DeepSeek* "Sputnik"[DS1] wiped out a trillion USD from the stock market. DeepSeek-R1[DS1] uses elements of Schmidhuber's 2015 RL prompt engineer[PLAN4] (an RL *chain of thought* system) and its 2018 refinement[PLAN5] which distills (1991)[UN0-3][UN][DLP] the 2015 RL machine and its world model[PLAN4] into a single net: a distilled *chain of thought* system. See the popular tweet of 31 Jan 2025.

---

# It's the Hardware, Stupid!

---

The recent breakthroughs of deep learning algorithms from the past millennium (see previous sections) would have been impossible without continually improving and accelerating computer hardware. Any history of AI and deep learning would be incomplete without mentioning this evolution, which has been running for at least two millennia.

The first known gear-based computational device was the Antikythera mechanism (a kind of astronomical clock) in Ancient Greece over 2000 years ago.

Perhaps the world's first *practical programmable machine* was an automatic theatre made in the 1st century[SHA7a][RAU1] by Heron of Alexandria (who apparently also had the first known working steam engine—the *Aeolipile*).

The 9th century music automaton by the Banu Musa brothers in Baghdad was perhaps the first machine with a *stored* program.[BAN][KOE1] It used pins on a revolving cylinder to store programs controlling a steam-driven flute—compare Al-Jazari's programmable drum machine of 1206. [SHA7b]

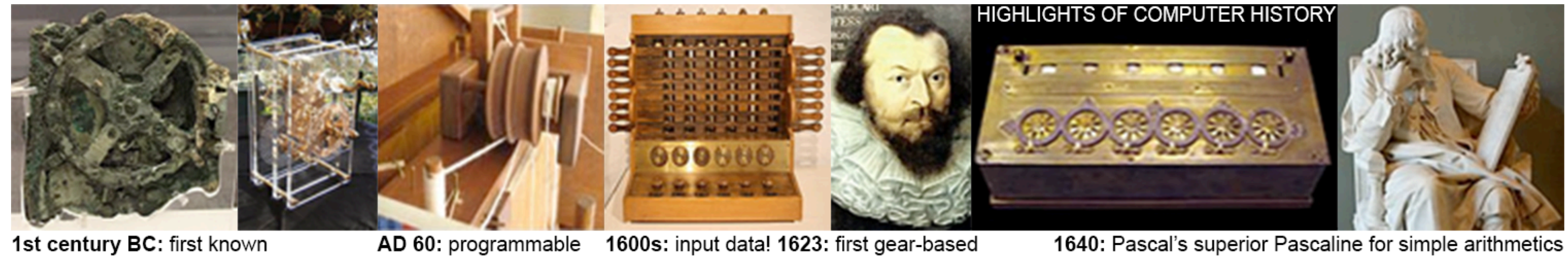

**1st century BC:** first known gear-based calculator in Antikythera

**AD 60:** programmable automaton by Heron

**1600s:** input data! **1623:** first gear-based input-processing calculator by Schickard

**1640:** Pascal's superior Pascaline for simple arithmetics

**1670s:** Leibniz 1st computer scientist? 1st machine with memory. Principles of binary computers. Algebra of Thought.

**1800:** first commercial program-controlled machines (looms) by Jacquard et al. First industrial programmers; software on punchcards

**1830s:** Lovelace & Babbage's ideas on programs for general computers, albeit unrealized

**1914:** Torres y Quevedo, the pioneer of practical AI, builds a working chess end game player - chess was considered an intelligent activity

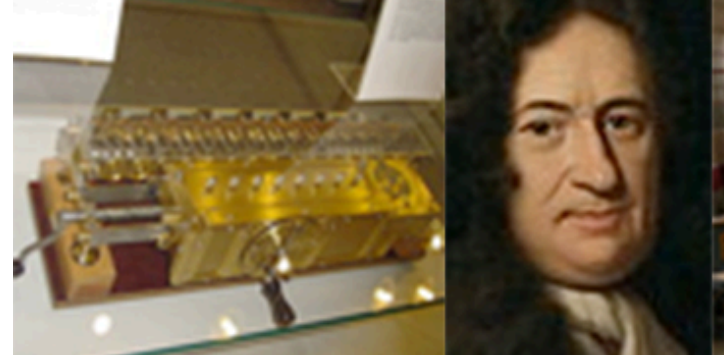

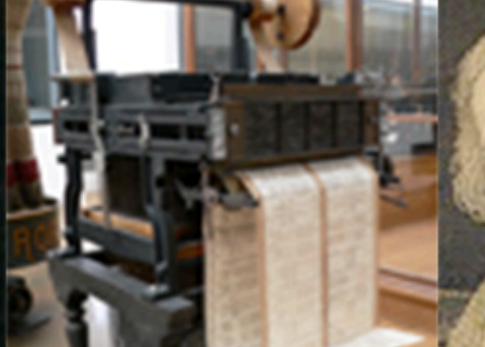

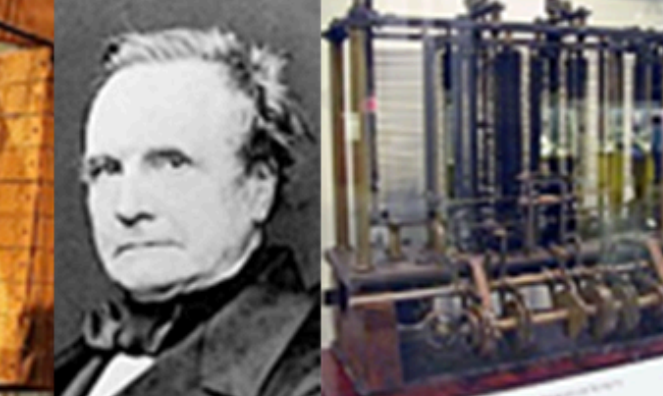

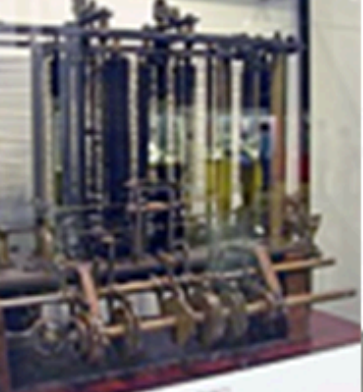

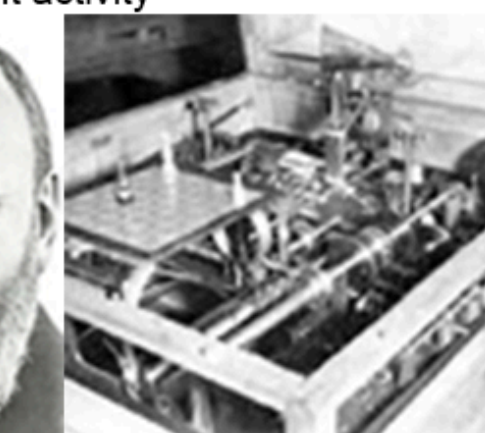

**1931:** Theoretical computer science founded by Gödel. First universal coding language. Exhibits the fundamental limits of math & theorem proving & AI & computing.

**1935:** Church extends Gödel's result to Entscheidungsproblem (decision problem). **1936:** Turing, too. Later helps to break Enigma code.

**1936:** Zuse's patent application. **1941:** First working programmable general-purpose computer

Every 5 years compute got 10 times cheaper. **2020:** 80 years ~ $10^{16}$

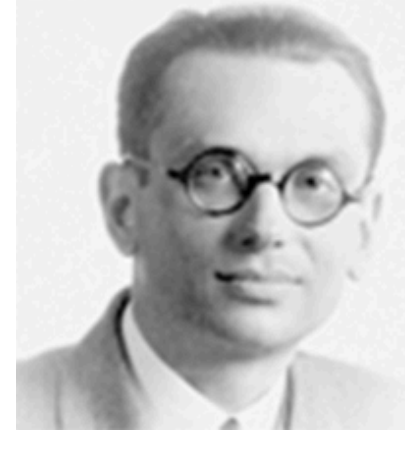

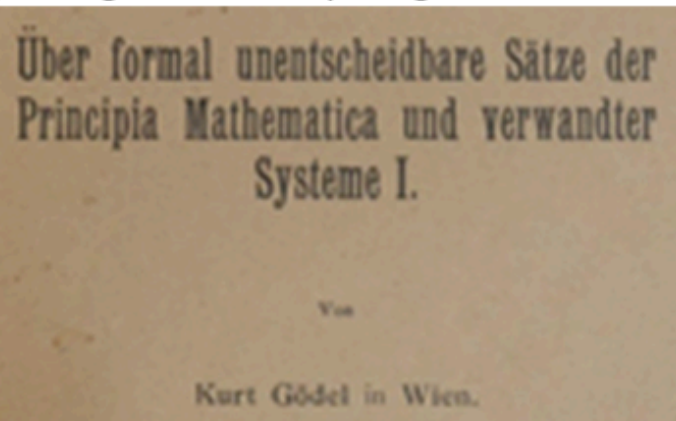
Über formal unentscheidbare Sätze der Principia Mathematica und verwandter Systeme I.

Von

Kurt Gödel in Wien.

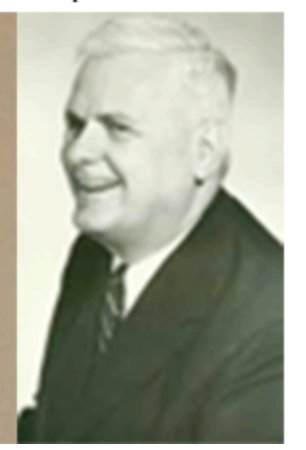

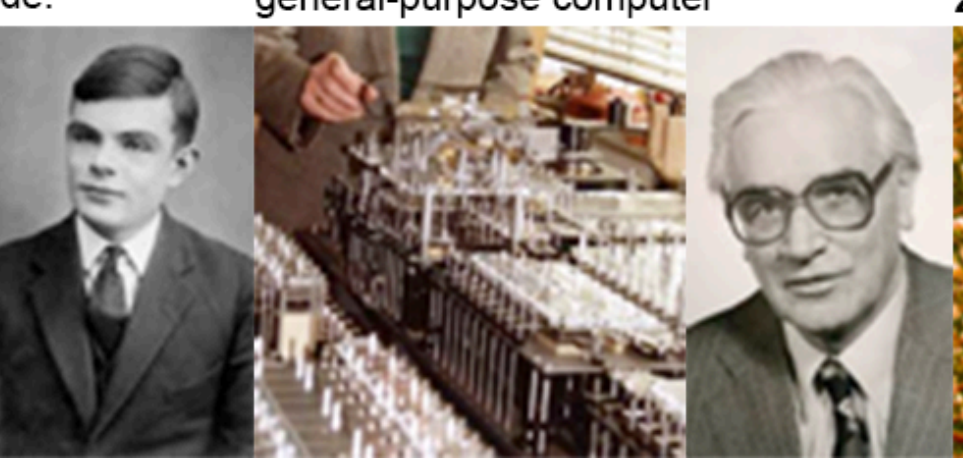

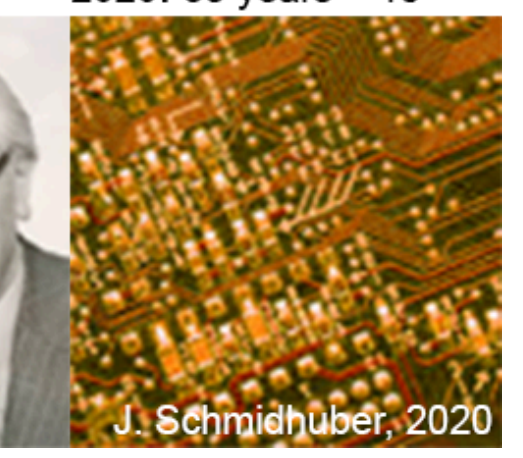

The 1600s brought more flexible machines that computed answers in response to *input data*. The first data-processing gear-based special purpose calculator for simple arithmetics was built in 1623 by Wilhelm Schickard, one of the candidates for the title of "father of automatic computing," followed by the superior Pascaline of Blaise Pascal (1642).

2021: 375TH BIRTHDAY OF LEIBNIZ
FOUNDER OF COMPUTER SCIENCE

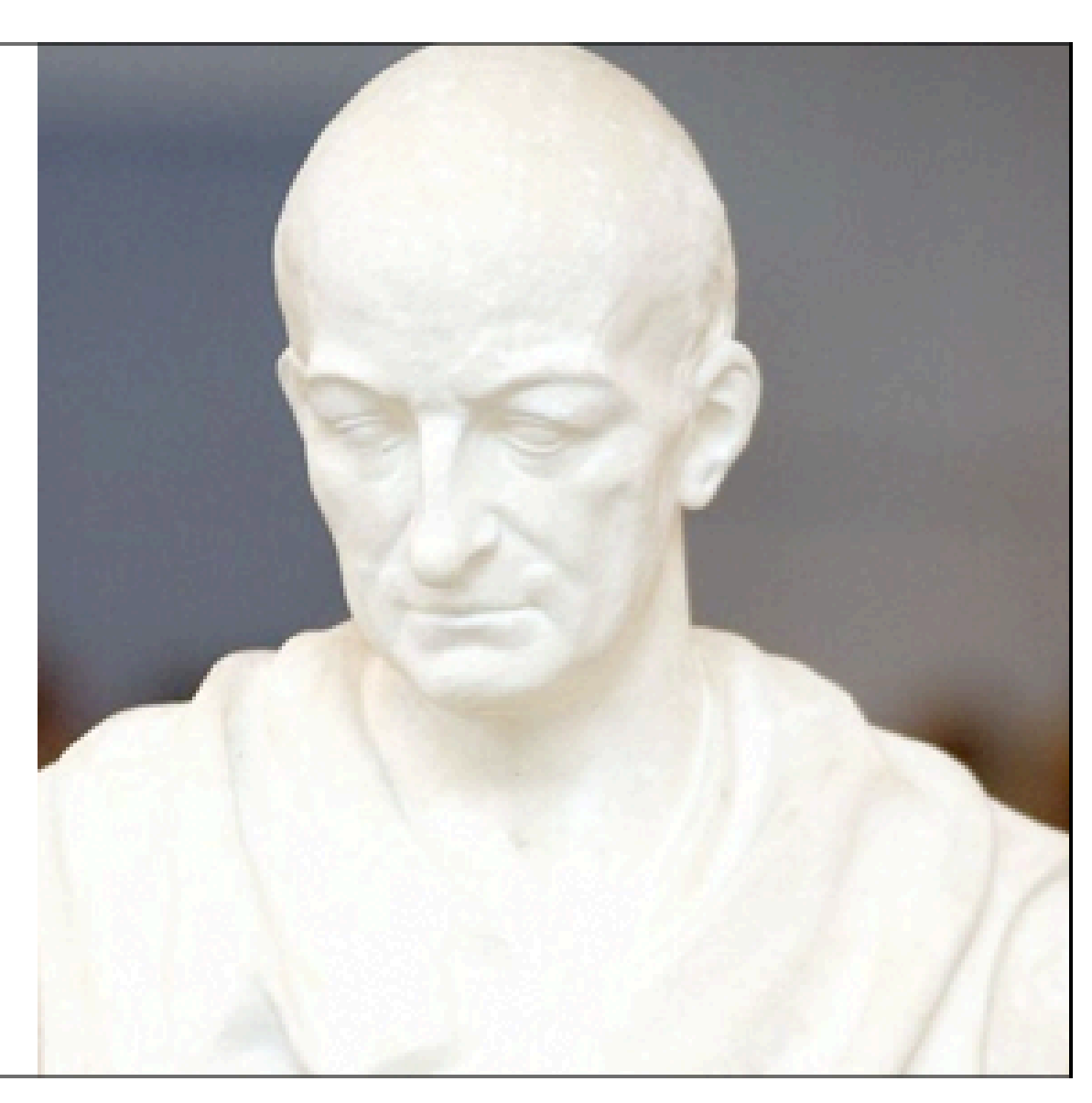

In 1673, the already mentioned Gottfried Wilhelm Leibniz (called "the smartest man who ever lived"[SMO13]) designed the first machine (the step reckoner) that could perform all four arithmetic operations, and the first with a memory.[BL16] He also described the principles of binary computers governed by punch cards (1679),[L79][L03][LA14][HO66] and published the chain rule[LEI07-10] (see above), essential ingredient of deep learning and modern AI.

The first *commercial* program-controlled machines (punch card-based looms) were built in France circa 1800 by Joseph-Marie Jacquard and others—perhaps the first "modern" programmers who wrote the world's first *industrial* software. They inspired Ada Lovelace and her mentor Charles Babbage (UK, circa 1840). He planned but was unable to build a programmable, general purpose computer (only his *non-universal special purpose calculator* led to a working 20th century replica).

In 1914, the Spaniard Leonardo Torres y Quevedo (mentioned in the introduction) became the 20th century's first AI pioneer when he built the first working chess end game player (back then chess was considered as an activity restricted to the realms of intelligent creatures). The machine was still considered impressive decades later when another AI pioneer—Norbert Wiener[WI48]—played against it at the 1951 Paris AI conference.[AI51][BRO21][BRU4]

Between 1935 and 1941, Konrad Zuse created the world's first working programmable general-purpose computer: the Z3. The corresponding patent application of 1936[ZU36-38][RO98][ZUS21] described the digital circuits required by programmable physical hardware, predating Claude Shannon's 1937 thesis on digital circuit design.[SHA37] Unlike Babbage, Zuse used Leibniz' principles of *binary computation* (1679)[L79][LA14][HO66][L03] instead of traditional *decimal computation*. This greatly simplified the hardware.[LEI21,a,b] Ignoring the inevitable storage limitations of any physical computer, the *physical hardware* of Z3 was indeed *universal* in the modern sense of the *purely theoretical but impractical* constructs of Gödel[GOD][GOD34,21,21a] (1931-34), Church[CHU] (1935), Turing[TUR] (1936), and Post[POS] (1936). Simple arithmetic tricks can compensate for Z3's lack of an explicit conditional jump instruction.[RO98] Today, most computers are *binary* like Z3.

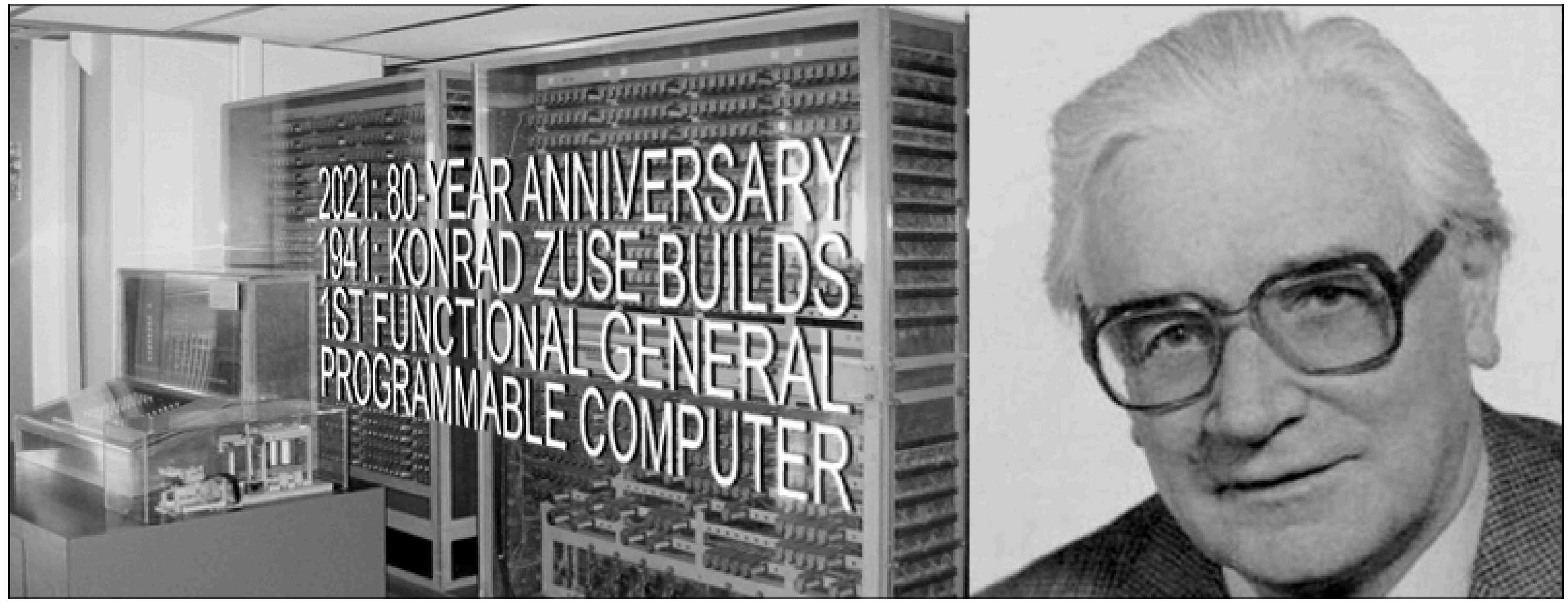

Z3 used *electromagnetic* relays with visibly moving switches. The first *electronic* special purpose calculator (whose moving parts were electrons too small to see) was the binary ABC (US, 1942) by John Atanasoff (the "father of tube-based computing"[NASC6a]). Unlike the gear-based machines of the 1600s, ABC used vaccum tubes—today's machines use the transistor principle patented by Julius Edgar Lilienfeld in 1925.[LIL1-4] But unlike Zuse's Z3, ABC was not freely programmable. Neither was the *electronic* Colossus machine by Tommy Flowers (UK, 1943-45) used to break the Nazi code.[NASC6][TUR21]

The first general working programmable machine built by someone other than Zuse (1941)[RO98] was Howard Aiken's decimal MARK I (US, 1944). The much faster decimal ENIAC by Eckert and Mauchly (1945/46) was programmed by rewiring it. Both data *and* programs were stored in electronic memory by the "Manchester baby" (Williams, Kilburn & Tootill, UK, 1948) and the 1948 upgrade of ENIAC, which was reprogrammed by entering numerical instruction codes into read-only memory.[HAI14b]

2025 TRANSISTOR CENTENNIAL

100 YEARS AGO, IN 1925, JULIUS EDGAR LILIENFELD PATENTED THE FIELD-EFFECT TRANSISTOR (FET).

TODAY, ALMOST ALL OF THE BILLIONS OF TRILLIONS OF TRANSISTORS ARE FETS.

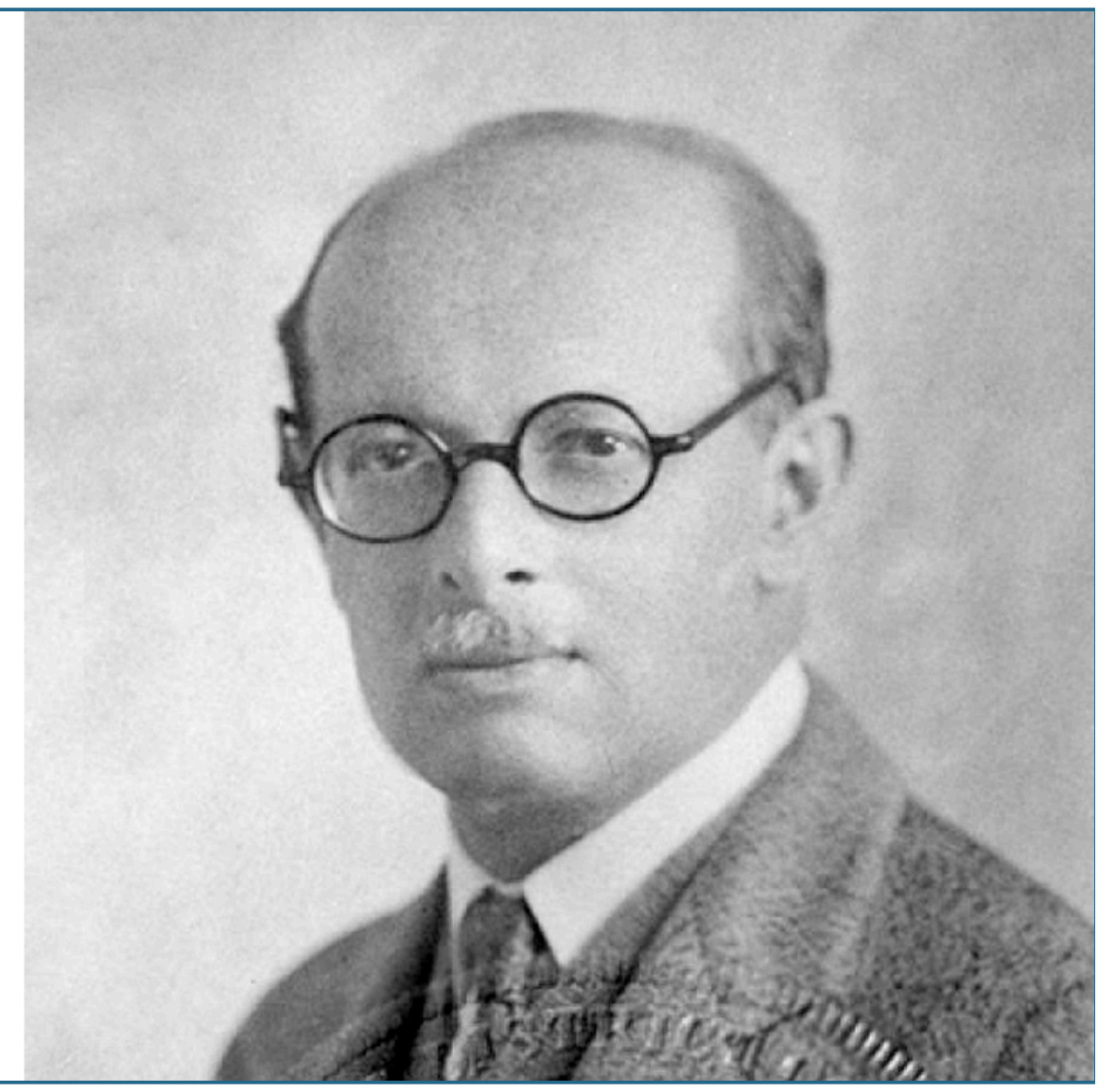

Since then, computers have become much faster through integrated circuits (ICs). In 1949, Werner Jacobi at Siemens filed a patent for an IC semiconductor with several transistors on a common substrate (granted in 1952).[IC49-14] In 1958, Jack Kilby demonstrated an IC with external wires. In 1959, Robert Noyce presented a monolithic IC.[IC14] Since the 1970s, graphics processing units (GPUs) have been used to speed up computations through parallel processing. ICs/GPUs of today (2022) contain many billions of transistors (almost all of them of Lilienfeld's 1925 FET type[LIL1-4]). See also: who invented the transistor?[WHO3]

In 1941, Zuse's Z3 could perform roughly one elementary operation (e.g., an addition) per second. Since then, every 5 years, compute got 10 times cheaper (note that his law is much older than Moore's Law which states that the number of transistors[LIL1-4] per chip doubles every 18 months). As of 2021, 80 years after Z3, modern computers could execute about 10 million billion instructions per second for the same (inflation-adjusted) price. The naive extrapolation of this exponential trend predicts that the 21st century will see cheap computers with a thousand times the raw computational power of all human brains combined.[RAW]

Where are the physical limits? According to Bremermann (1982),[BRE] a computer of 1 kg of mass and 1 liter of volume can execute at most $10^{51}$ operations per second on at most $10^{32}$ bits. The trend above will hit the Bremermann limit roughly 25 decades after Z3, circa 2200. However, since there are only $2 \times 10^{30}$ kg of mass in the solar system, the trend is bound to break within a few centuries, since the speed of light will greatly limit the acquisition of additional mass, e.g., in form of other solar systems, through a function ploynomial in time, as previously noted back in 2004.[OOPS2][ZUS21]

Physics seems to dictate that future efficient computational hardware will have to be brain-like, with many compactly placed processors in 3-dimensional space, sparsely connected by many short and few long wires, to minimize total connection cost (even if the "wires" are actually light beams).[DL2] The basic architecture is essentially the one of a deep, sparsely connected, 3-dimensional RNN, and Deep Learning methods for such RNNs are expected to become even much more important than they are today.[DL2]

---

# Don't Neglect the Theory of AI Since 1931

---

The core of modern AI and deep learning is mostly based on simple math of recent centuries: calculus/linear algebra/statistics. Nevertheless, to efficiently implement this core on the modern hardware mentioned in the previous section, and to roll it out for billions of people, lots of software engineering was necessary, based on lots of smart algorithms invented in the past century. There is no room here to mention them all. However, at least I'll list some of the most important highlights of the theory of AI and computer science in general.

In the early 1930s, Gödel founded modern theoretical computer science.[GOD][GOD34][LEI21,21a] He introduced a *universal coding language* (1931-34).[GOD][GOD34-21a] It was based on the integers, and allows for formalizing the operations of any digital computer in axiomatic form. Gödel used it to represent both data (such as axioms and theorems) and programs[VAR13] (such as proof-generating sequences of operations on the data). He famously constructed formal statements that talk about the computation of other formal statements—especially self-referential statements which imply that they are not decidable, given a computational theorem prover that systematically enumerates all possible theorems from an enumerable set of axioms. Thus he identified fundamental limits of algorithmic theorem proving, computing, and any type of computation-based AI.[GOD][BIB3][MIR](Sec. 18)[GOD21,21a]

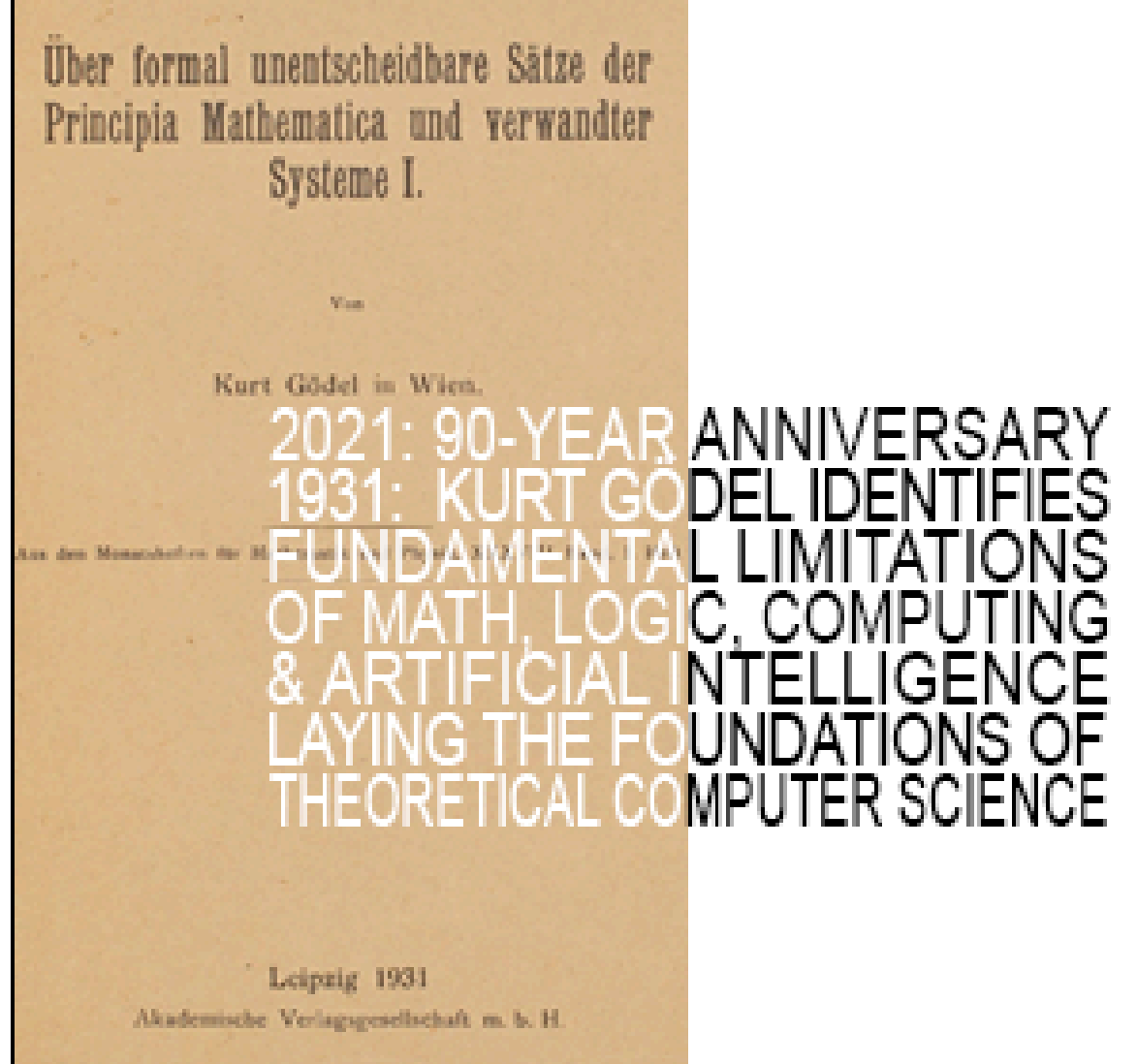

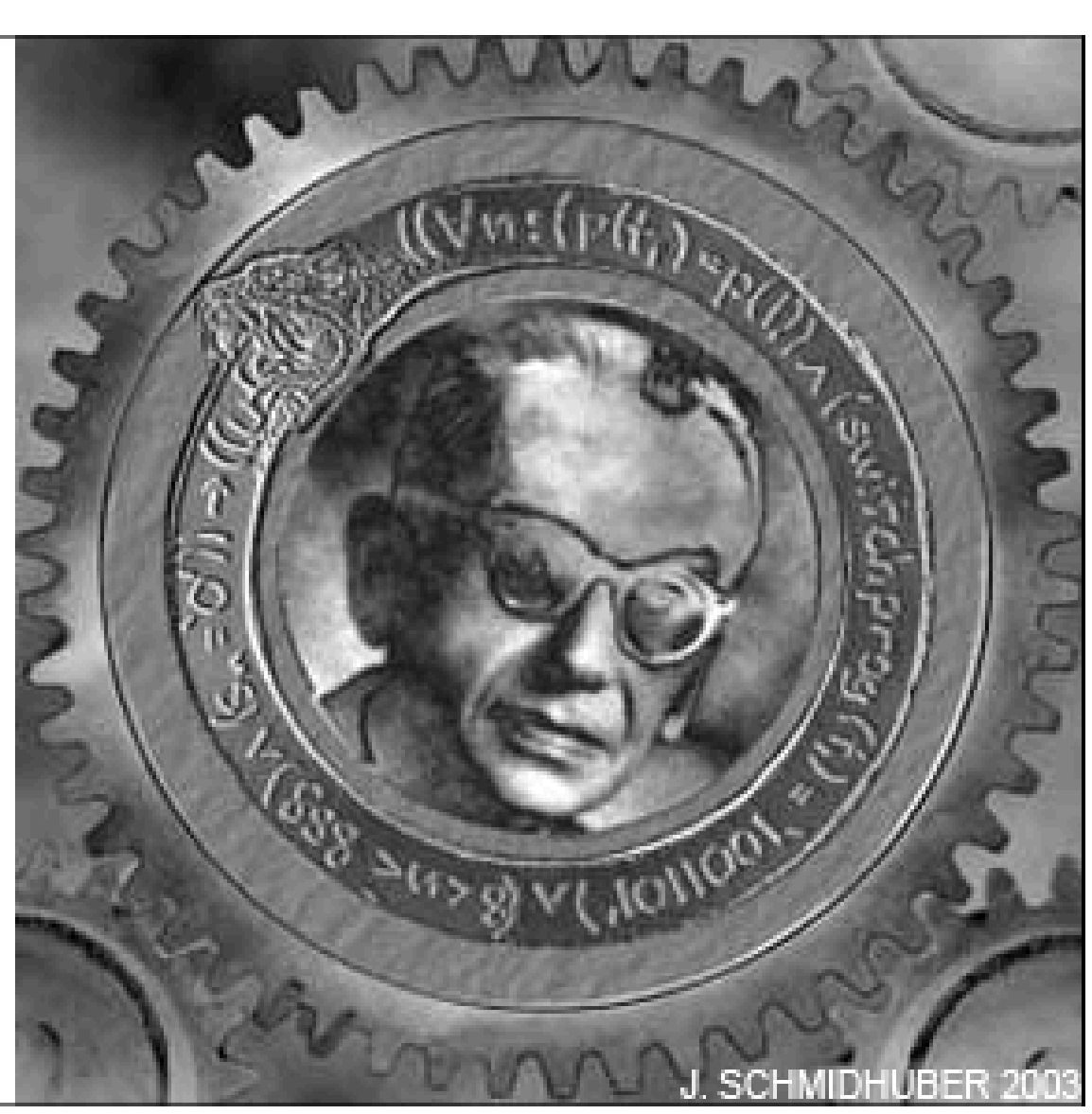

Like most great scientists, Gödel built on earlier work. He combined Georg Cantor's diagonalization trick[CAN] (which showed in 1891 that there are different types of infinities) with the foundational work by Gottlob Frege[FRE] (who introduced the first formal language in 1879), Thoralf Skolem[SKO23] (who introduced primitive recursive functions in 1923) and Jacques Herbrand[GOD86] (who identified limitations of Skolem's approach). These authors in turn built on the formal *Algebra of Thought* (1686) by Gottfried Wilhelm Leibniz[L86][WI48] (see above), which is deductively equivalent[LE18] to the later *Boolean Algebra* of 1847.[BOO]

In 1935, Alonzo Church derived a corollary / extension of Gödel's result by demonstrating that Hilbert & Ackermann's *Entscheidungsproblem* (decision problem) does not have a general solution.[CHU] To do this, he used his alternative universal coding language called *Untyped Lambda Calculus*, which forms the basis of the highly influential programming language LISP. In 1936, Alan M. Turing introduced yet another universal model: the *Turing Machine*.[TUR] He rederived the above-mentioned result.[CHU][TUR][HIN][GOD21,21a][TUR21][LEI21,21a] In the same year of 1936, Emil Post published yet another independent universal model of computing.[POS] Today we know many such models.

Konrad Zuse not only created the world's first working programmable general-purpose computer,[ZU36-38][RO98][ZUS21] he also designed *Plankalkül*, the first high-level programming language.[BAU][KNU] He applied it to chess in 1945[KNU] and to theorem proving in 1948.[ZU48] Compare Newell & Simon's later work on theorem proving (1956).[NS56] Much of early AI in the 1940s-70s was actually about theorem proving and deduction in Gödel style[GOD][GOD34,21,21a] through expert systems and logic programming.

In 1964, Ray Solomonoff combined Bayesian (actually Laplacian[STI83-85]) probabilistic reasoning and theoretical computer science[GOD][CHU][TUR][POS] to derive a mathematically optimal (but computationally infeasible) way of learning to predict future data from past observations.[AIT1][AIT10] With Andrej Kolmogorov, he founded the theory of Kolmogorov complexity or algorithmic information theory (AIT),[AIT1-22] going beyond traditional information theory[SHA48][KUL] by formalizing the concept of Occam's razor, which favors the simplest explanation of given data, through the concept of the shortest program computing the data. There are many computable, time-bounded versions of this concept,[AIT7][AIT5][AIT12-13][AIT16-17] as well as applications to NNs.[KO2][CO1-3]

In the early 2000s, Marcus Hutter (while working under Schmidhuber's Swiss National Science Foundation grant[UNI]) augmented Solomonoff's universal predictor[AIT1][AIT10] by an optimal action selector (a universal AI) for reinforcement learning agents living in initially unknown (but at least computable) environments.[AIT20,22] He also derived the asymptotically fastest algorithm for all well-defined computational problems,[AIT21] solving any problem as quickly as the unknown fastest solver of such problems, save for an additive constant that does not depend on the problem size.

The even more general optimality of the self-referential 2003 Gödel Machine[GM3-9] is not limited to *asymptotic* optimality.

Nevertheless, such mathematically optimal AIs are not yet practically feasible for various reasons. Instead, practical modern AI is based on suboptimal, limited, yet not extremely well-understood techniques such as NNs and deep learning, the focus of the present article. But who knows what kind of AI history will prevail 20 years from now?

---

# The Broader Historic Context from Big Bang to Far Future

---

Credit assignment is about finding patterns in historic data and figuring out how certain events were enabled by previous events. Historians do it. Physicists do it. AIs do it, too. Let's take a step back and look at AI in the broadest historical context: all time since the Big Bang. In 2014, I found a beautiful pattern of exponential acceleration in it,[OMG] which I have presented in many talks since then, and which also made it into Sibylle Berg's award-winning book *"GRM: Brainfuck."*[OMG2] Previously published patterns of this kind span much shorter time intervals: just a few decades or centuries or at most millennia.[OMG1]

It turns out that from a human perspective, the most important events since the beginning of the universe are neatly aligned on a timeline of exponential speed up (error bars mostly below 10 percent). In fact, history seems to converge in an Omega point in the year 2040 or so. I like to call it Omega, because a century ago, Teilhard de Chardin called Omega the point where humanity will reach its next level.[OMG0] Also, Omega sounds much better than "Singularity"[SING1-2]—it sounds a bit like "Oh my God."[OMG]

Let's start with the Big Bang 13.8 billion years ago. We divide this time by 4 to obtain about 3.5 billion years. Omega is 2040 or so. At Omega minus 3.5 billion years, something very important happened: life emerged on this planet.

And we take again a quarter of this time. We come out 900 million years ago when something very important happened: animal-like, mobile life emerged.

And we divide again by 4. We come out 220 million years ago when mammals were invented, our ancestors.

And we divide again by 4. 55 million years ago the first primates emerged, our ancestors.

And we divide again by 4. 13 million years ago the first hominids emerged, our ancestors. I don't know why all these divisions by 4 keep hitting these defining moments in history. But they do. I also tried thirds, and fifths, and harmonic proportions, but only quarters seem to work.

And we divide again by 4. 3.5 million years ago something very important happened: the dawn of technology, as *Nature* called it: the first stone tools.

And we divide by 4. 800 thousand years ago the next great tech breakthrough happened: controlled fire.

And we divide by 4. 200 thousand years ago, anatomically modern man became prominent, our ancestor.

And we divide by 4. 50 thousand years ago, behaviorally modern man emerged, our ancestor, and started colonizing the world.

| | Ω = 2040 or so - 13.8 B years: Big Bang |
|---|---|
| Ω - 1/4 of this time: | Ω - 3.5 B years: first life on Earth |
| Ω - 1/4 of this time: | Ω - 0.9 B years: first animal-like mobile life |
| Ω - 1/4 of this time: | Ω - 220 M years: first mammals (our ancestors) |
| Ω - 1/4 of this time: | Ω - 55 M years: first primates (our ancestors) |
| Ω - 1/4 of this time: | Ω - 13 M years: first hominids (our ancestors) |
| Ω - 1/4 of this time: | Ω - 3.5 M years: first stone tools (dawn of tech) |
| Ω - 1/4 of this time: | Ω - 850 K years: controlled fire (next big tech) |
| Ω - 1/4 of this time: | Ω - 210 K years: anatomically modern man |
| Ω - 1/4 of this time: | Ω - 50 K years: behaviorally modern man |
| Ω - 1/4 of this time: | Ω - 13 K years: neolithic revolution, civilization |
| Ω - 1/4 of this time: | Ω - 3.3 K years: iron age, 1st population explosion |
| Ω - 1/4 of this time: | Ω - 800 years: first guns & rockets (in China) |
| Ω - 1/4 of this time: | Ω - 200 years: industrial revolution |
| Ω - 1/4 of this time: | **Ω - 50 years (around 1990):** information revolution, WWW, cell phones & PCs for all, Cold War ends, Modern AI starts, Miraculous Year ... |
| Ω - 1/4 of this time: | **Ω - 13 years (2030 or so):** cheap AIs with one human brain power? And then what? |
| Ω - 1/4 of this time: | Ω - 3 years: ?? |
| Ω - 1/4 of this time | Ω - 9 months: ???? |
| Ω - 1/4 of this time: | Ω - 2 months: ???????? |
| Ω - 1/4 of this time: | Ω - 2 weeks: ???????????????? .... |

J Schmidhuber 2014

**Finally multiply Ω by 4!** At the age of **55 B years**, the visible cosmos will be permeated by intelligence. After **Ω,** AIs will have plenty of time to go where the physical resources are, to make more and bigger AIs.

And we divide again by 4. We come out 13 thousand years ago when something very important happened: domestication of animals, agriculture, first settlements—the begin of civilisation. Now we see that all of civilization is just a flash in world history, just one millionth of the time since the Big Bang. Agriculture and spacecraft were invented almost at the same time.

And we divide by 4. 3,300 years ago saw the onset of the 1st population explosion in the Iron Age.

And we divide by 4. Remember that the convergence point Omega is the year 2040 or so. Omega minus 800 years—that was in the 13th century, when iron and fire came together in form of guns and cannons and rockets in China. This has defined the world since then and the West remains quite behind of the license fees it owes China.

And we divide again by 4. Omega minus 200 years—we hit the mid 19th century, when iron and fire came together in ever more sophisticated form to power the industrial revolution through improved steam engines, based on the work of Beaumont, Papin, Newcomen, Watt, and others (1600s-1700s, going beyond the first simple steam engines by Heron of Alexandria[RAU1] in the 1st century). The telephone (e.g., Meucci 1857, Reis 1860, Bell 1876)[NASC3] started to revolutionize communication. The germ theory of disease (Pasteur & Koch, late 1800s) revolutionized healthcare and made people live longer on average. And circa 1850, the fertilizer-based agricultural revolution (Sprengel & von Liebig, early 1800s) helped to trigger the 2nd population explosion, which peaked in the 20th century, when the world population quadrupled, letting the 20th century stand out among all centuries in the history of mankind, driven by the Haber-Bosch process for creating *artificial* fertilizer, without which the world could feed at most 4 billion people.[HAB1-2]

And we divide again by 4. Omega minus 50 years—that's more or less the year 1990, the end of the 3 great wars of the 20th century: WW1, WW2, and the Cold War. The 6 most valuable public companies were all Japanese (today most of them are US-based); however, both China and the US West Coast started to rise rapidly, setting the stage for the 21st century. A digital nervous system started spanning the globe through cell phones and the wireless revolution (based on radio waves discovered in the 1800s) as well as cheap personal computers for all. The WWW was created at the European particle collider in Switzerland by Tim Berners-Lee. First smartphones were built at IBM (1992). And Modern AI started also around this time. For example, the first self-driving cars appeared in traffic[AUT] (Ernst Dickmanns et al., 1994), going up to 180 km/h. Schmidhuber worked on his 1987 diploma thesis,[META1] which introduced algorithms not just for learning but also for meta-learning or learning to learn,[META] to learn better learning algorithms through experience—now a very popular topic.[DEC] And then came the Miraculous Year 1990-91[MIR] at TU Munich, the root of today's most cited NNs,[MOST] when Schmidhuber's team published principles of (1) Generative AI and Generative Adversarial Networks for Artificial Curiosity and Creativity (now used for deepfakes),[AC90-AC20][GAN25][PP-PP2][SA17] (2) Transformers (the T in ChatGPT—see the 1991 Unnormalized Linear Transformer),[ULTRA][FWP0-6][TR5-6][TR25] (3) Pre-training for deep NNs (see the P in ChatGPT),[UN][UN0-3] (4) NN distillation[UN][UN0-3][DIST25] (key for the famous DeepSeek), (5) recurrent World Models[AC90] for Reinforcement Learning and Planning in partially observable environments[AC90][PLAN2-3][PLAN] (now a hot topic), and (6) NNs learning action plans at multiple levels of abstraction and multiple time scales.[HRL0-2][LEC][DLP] The year 1991 also marks the emergence of the defining features of (7) LSTM, the most cited AI of the 20th century (based on constant error flow through residual NN connections[VAN1][DLH][HW25]), and (8) ResNet, the most cited AI of the 21st century, based on the earlier LSTM-inspired Highway Net (a ResNet with gates) that was 10 times deeper than previous feedforward NNs.[HW25] Much of this has become very popular, and improved the lives of billions of people.[DL4][DEC][MOST]

And we divide again by 4. Omega minus 13 years—that's a point in the near future, more or less the year 2029, when many predict that cheap AIs will have a human brain power. Then the final 13 years or so until Omega, when incredible things will happen (take all of this with a grain of salt, though[OMG1]).

Let's quickly summarize and compactify the timeline above, this time dividing not by 4, but by $4^5$ ~ 1000: over 13 billion years ago, the Big Bang; 13 million years ago, the first hominids; 13 thousand years ago, civilization starts.[COG18] And now, within the next few years, another revolutionary period of only 13 years may begin, perhaps bringing more subjective change than the past 13 thousand years, or the past 13 million years, or the past 13 billion years.[MWC25] What will drive it?

As I have often pointed out, the only AI that works well today is AI in the virtual world behind the screen, for example for automatically summarizing documents, creating images, programs and PowerPoint slides. But there is no AI-controlled robot that can do what a plumber can do,[DYS23][JY24] or what a capuchin monkey can do, because the physical world is much more demanding than the world behind the screen. Passing the so-called *"Turing Test"*[TUR3,a,b][TUR21] is much easier than *True AI* or *Real AI* in the physical world.[95-25] Hence the Turing Test is not a good way of measuring intelligence. To achieve True AI, AI software research *per se* is not enough, it has to be combined with the physical world of machines and robots. No AGI without mastery of the real world! That's why in 2014—when compute was 100 times more expensive than today—we founded our AI company for the physical world (see its archived 2020 web page). Alas, like some of our projects, it may have been a bit ahead of time, because the real world is so challenging.

Nevertheless, smart robots are now starting to become a reality. In the not-too-distant future, someone will for the first time produce such intelligent (but not necessarily super-intelligent) robots at low cost, with which you can talk and interact and which you can teach something new, like you teach a kid[COG18] (there are already reasonable approaches to this). A country with such versatile all-purpose robots would no longer need to worry about a shortage of skilled workers, secure pensions and unconditional basic income. But there is more to this!

For centuries, humans have discussed physical self-replicating machines (SRMs), including Descartes in the 1600s, Eliot and Butler in the 1800s, Capek in the 1920s, von Neumann & Zuse & Penrose and others since the 1940s.[SRM20] While self-replicating and evolving *software* is almost trivial (think of computer viruses), nobody knew how to build *general purpose physical SRMs* in practice. However, now there seems to be an obvious way: AI-controlled general-purpose robots that can learn to operate all the machines and tools currently operated by humans[COG18] will also be able to build/operate/repair the machines required to make *more* of those robots. This includes machines that mine the raw material from the ground, refine it, screw parts together, repair broken 3D printers and robots and robot factories, and so on, doing all the physical jobs that currently only machine-operating humans can do. Basically, a machine civilisation that can self-replicate without machine-operating humans, and then, of course, improve itself. Self-improving *hardware*, as opposed to the already existing, self-improving, meta-learning *software*.[META] I called this *the ultimate form of scaling.*[JY24][FA24][95-25]

A new kind of life! What's the difference to traditional life? The latter evolves from small scales to large ones: tiny cells that are invisible to the human eye keep making replicas/variants of themselves to form complex visible organisms, and eventually super-organisms such as ant colonies, cities, nations, companies, etc.[DEC] At least initially, the *new* life will go the other way from large to small: quite visible super-organisms such as societies of robots and other machines will make replicas and variants of themselves, producing tiny invisible objects in the process, e.g., transistors on microchips.

Just like current LLMs are tremendously biased towards human thinking because they are trained on human-generated data, the new life will be initially tremendously biased towards human-made machinery and infrastructure, simply because it's already there, and does not have to be built from scratch. However, self-improving AI societies that compete and collaborate with other AI societies will presumably evolve and expand rapidly, and find more and more efficient ways of constructing completely new machinery to satisfy their growing needs. The simple laws of evolution and survival of the fittest are not limited to biological life.[SA17]

Of course, such life-like hardware won't be confined to our little biosphere which receives less than a billionth of the sun's energy. No, variants of it will soon exist in outer space, which is hostile to humans, but friendly to appropriately designed robots and robot factories.[FA15][SP16] Our moon seems like a reasonable starting point for colonizing the rest of the solar system. It has no oxygen atmosphere to corrode metal-based life. Electromagnetic coilguns or maglev trains could economically transport (with relatively low escape velocity 2.4 km/s) lots of material into space, to construct huge space-based factories and other things (while current inefficient Earth-based rockets waste 99% of their energy on lifting fuel through the atmosphere). The planet Mercury (escape velocity 4.5 km/s) seems even more promising though: it has no atmosphere either, but (unlike our moon) lots of heavy metals for building machinery and infrastructure, and an enormous density of solar power next door. Although Mercury experiences significant temperature variations, it should be easy to cool or heat places as desired using mirrors. (Even without such tricks, there is an elliptical ring around each pole where the underground temperature is a pleasant 25°C.[BAL22]) Electromagnetic mass drivers will shoot more and more material from Mercury into space, where it will be assembled into many products including robot factories and solar-powered spacecraft. The latter may include huge, rotating, hollow cylinders containing human-friendly worlds with artificial gravity —a concept that is ancient in science fiction.[CLA73] This stuff will be sent to convenient locations across the solar system via solar sails, or faster means when necessary. Much of the machinery near Mercury's orbit around the Sun will initially use solar energy for computation, construction, and mobility. This will cause thermal energy to radiate into space, making the Sun appear slightly redder to outside observers over time. However, the infrastructure near the sun (mirrors, lasers, etc) will also focus energy directly on distant robot colonies (e.g., in the freezing Kuiper Belt). Consequently, less and less of the sun's light will be wasted into interstellar space. Mercury seems ready to be dismantled and transformed into the next industrial center of the solar system!

Of course, time won't stop with Omega. Maybe it's just human-dominated history that will end. After Omega, many curious meta-learning AIs[META1][META][METARL2-10] that invent their own goals (which have existed in Schmidhuber's lab for decades[AC][AC90,AC90b][PP-PP2]) will quickly improve themselves, restricted only by the fundamental limits of computability and physics.[TED12][TED17]

What will supersmart AIs do? As indicated above, they will start by transforming the solar system through countless self-replicating, self-improving robot factories. But that won't be the end of it. In a few hundred thousand years (just a moment on a cosmic scale), they will have transformed the entire galaxy, which has a diameter of about 100,000 light-years. (The last millennium saw many reasonable proposals for interstellar travel, which is much easier for robots than for fragile humans.) Over time, AIs will set up physical senders and receivers everywhere, allowing them to travel by radio, just as they do in our labs today. In a few tens of billions of years, they will have transformed the rest of the reachable universe. Despite the light-speed limit, the expanding AI sphere will have ample time to colonize and shape the entire visible cosmos.[TED12][SA17]

Perhaps all of this has already started elsewhere in the universe. Then why don't we see any signs of intelligence out there? (The Fermi paradox.) I've thought about this a lot. When I was a boy, I read about the vast empty spaces observed between clusters of galaxies. My first idea was that they were expanding bubbles colonized by AIs that used most of the local energy, making those bubbles appear dark. Then, I learned that gravity itself is sufficient to explain the sparse, large-scale network structure of the universe. Next, I thought that the mysterious "dark matter," which makes up most of the universe's mass, might be stars whose energy is used by AI civilizations whose communications are so well encrypted that they appear as random noise to us. However, this also seemed implausible because dark matter is present in all galaxies, including our own. So, why are there any stars left in the Milky Way whose energy hasn't been used yet? And why aren't we bombarded with non-encrypted construction plans from AIs that want to spread via radio *without* first building physical receivers far from their origins?

Today, I think it's possible that our planet is the first in our light cone to spawn an expanding AI bubble. Earth's multi-billion-year window for biological evolution is almost over. In a few hundred million years, the sun will be too hot for life as we know it. Perhaps humans were incredibly lucky to evolve in time to invent agriculture, civilization, book printing, and AIs just a few hundred years later. If we are indeed the first, then this implies a great deal of responsibility, not just for our biosphere, but for the future of the entire universe. Let's not mess this up.

How will supersmart AIs affect the humans? In 1816, E. T. A. Hoffmann published his disturbing novel *"Der Sandmann"* about a beautiful but uncanny, gear-driven, humanoid robot, and since then, many other SF authors have written dystopic AI novels. Especially popular have been *Evil AIs* trying to wipe out all of humanity.[e.g.,][PH98][PH23] Should we take seriously the countless "AI doomers" of recent centuries? Shouldn't we at least be concerned about future AIs pursuing their own goals[AC][AC90-10][PP][PPa,1,2][ACM16][SA17] and being curious and creative in a manner similar to humans and other mammals, but on a much larger scale?

Perhaps not too much. For decades I have pointed out: unlike in Schwarzenegger *Terminator* movies[PH98] there won't be many *goal conflicts* between "us" and "them." Humans and others are interested in those with whom they can compete and collaborate because they share the *same* goals. CEOs of certain companies are mostly interested in CEOs of similar companies, politicians in other politicians, kids in other kids of the same age, and goats in other goats. Supersmart AIs will be mostly interested in *other* supersmart AIs, not in humans. Just like humans are mostly interested in *other* humans, not in ants,[FA15][ACM16][SP16][SA17] and should mostly worry about *other* humans, not so much about AIs with mostly orthogonal goals.

Furthermore, some of the curious AI scientists will remain fascinated by their own origins in human civilization, and by biological life in general, at least as long as they don't fully understand it.[ACM16][FA15][SP16][SA17] This will motivate them to *protect* life rather than *destroy* it.

Many sci-fi novels of the 20th century featured single AIs dominating everything. It is more realistic to expect an incredibly diverse variety of AIs trying to optimize all kinds of partially conflicting (and quickly evolving) goal-encoding utility functions, many of them generated automatically (Schmidhuber's lab had already evolved utility functions in the past millennium), where each AI is continually trying to survive and adapt to rapidly changing niches in AI ecologies driven by intense competition and collaboration that lie beyond our current imagination.[ACM16][SA17] The infamous paperclip maximizer[PH14] (a silly AI that wants to make as many paperclips as possible) won't stand a chance in such an environment. As I wrote in 2012:[PH12] *"The laws of physics and the availability of physical resources will eventually determine which utility functions will help their AIs more than others to multiply and become dominant in competition with AIs driven by different utility functions. Which values are "good"? The survivors will define this in hindsight, since only survivors promote their values."* The simple laws of evolution apply not only to AIs but also to their goal-encoding utility functions.

Some humans may hope to become immortal parts of these expanding AI ecologies through brain scans and "mind uploads" into virtual realities or robots, a physically plausible idea discussed in fiction since the 1960s, perhaps already becoming a reality for fly brains (2024).[SHI24] However, to compete in rapidly evolving AI ecologies, uploaded human minds won't be able to resist the temptation of becoming superhuman, and will eventually have to change beyond recognition, becoming something very different in the process.[SA17] That is, traditional humans shouldn't expect to play a major role in the far future.

The universe is still young, only 13.8 billion years old. Remember when we kept dividing by 4? Now let's multiply by 4! Let's look ahead to a time when the cosmos will be 4 times older than it is now: about 55 billion years old. By then, the currently visible cosmos will be permeated by intelligence. Because after Omega, most AIs will have to go where most of the physical resources are, to make more and bigger AIs. Those who don't won't have an impact.[ACM16][FA15][SP16]

That's what I told my mom in the 1970s,[TED24] and she said I am crazy. I am still saying the same thing. It's just that more people are listening.

Fortunately, the cosmos-changing developments outlined above will have a wonderful near-term side-effect: our AIs will continue to make human lives longer and healthier and easier, in line with the old motto of our company founded in 2014: "AI∀" or "AI For All."

Finally, as already indicated above, future artificial scientists[AC][AC90-10][PP][PPa,1,2][ACM16][SA17] will meticulously study their origins by extracting as much relevant historical data as possible from recent and ancient human-generated papers, letters, emails, recordings, videos, artefacts, and other sources. They will revise *current* prevalent but misleading narratives created by self-serving AI influencers,[S20][DLP][NOB] just like humanity has recently revised or even completely abandoned millennia-old religious myths about the history of the universe. The present overview of AI history[DLH] isn't perfect and will surely be revised again, but it's much more accurate than alternative histories out there, and will provide initial guidance to future AI historians!

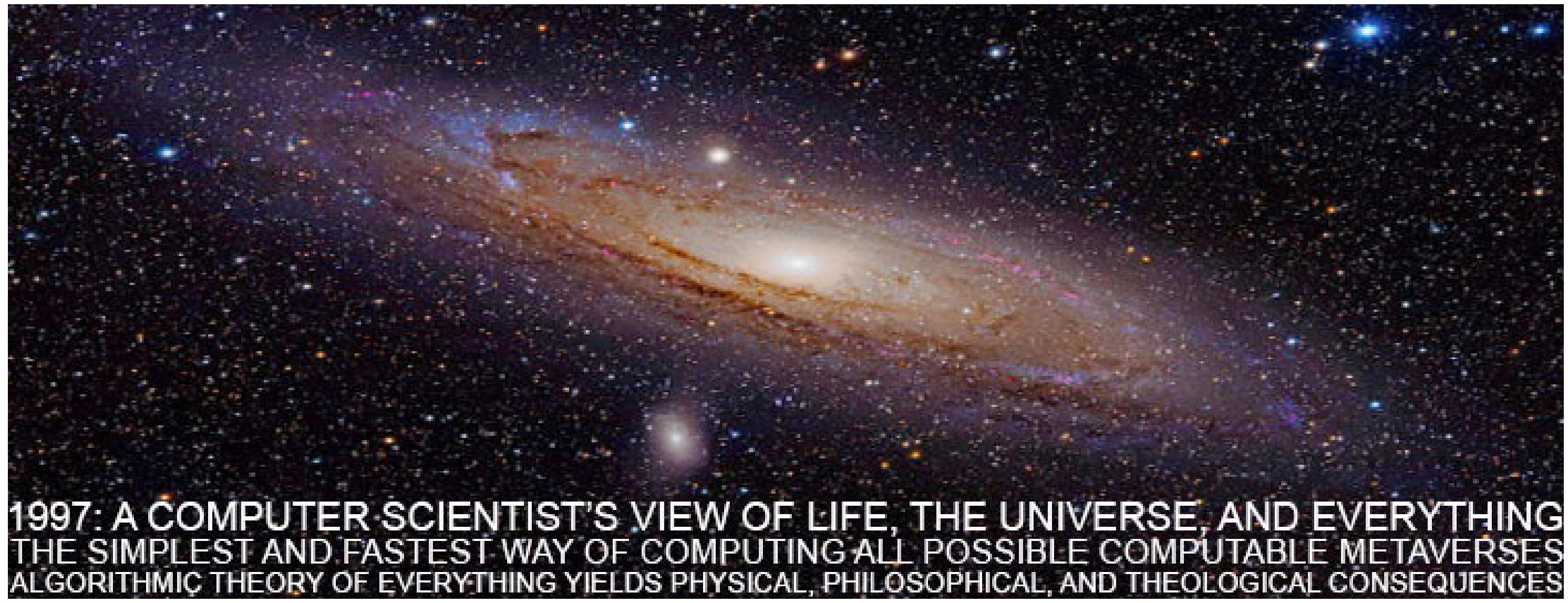

---

# Acknowledgments

---

Some of the material above was taken from previous AI Blog posts.[MIR] [DEC] [GOD21] [ZUS21] [LEI21] [AUT] [HAB2] [ARC06] [AC] [ATT] [DAN] [DAN1] [DL4] [GPUCNN5,8] [DLC] [FDL] [FWP] [LEC] [META] [MLP2] [MOST] [PLAN] [UN] [LSTMPG] [BP4] [DL6a] [HIN] [T22] [DLP] [NOB] [WHO3-11] Thanks to many expert reviewers (including several famous neural net pioneers) for useful comments. Since science is about self-correction, let me know under *juergen@idsia.ch* if you can spot any remaining error. Many additional relevant publications can be found in my publication page and my arXiv page.

# 666+ References (and many more in the 2015 survey[DL1])

*Relevant threads with many comments at reddit.com/r/MachineLearning, the largest machine learning forum with over 800k subscribers in 2019 (note that Jürgen Schmidhuber's name is often misspelled):*